\documentclass[11pt,onecolumn,fullpage]{article}

\usepackage{bm}
\usepackage{amsfonts}
\usepackage{amsmath}
\usepackage{amssymb}
\usepackage{cite}
\usepackage[amsmath,thmmarks]{ntheorem}

\usepackage{color, soul}
\usepackage{epic, eepic}
\usepackage[top=1in, bottom=1in, left=1in, right=1in]{geometry}
\usepackage{appendix}
\usepackage{footnote}
\usepackage{booktabs}
\usepackage[pdftex]{graphicx}
\usepackage[ruled,vlined]{algorithm2e}
\usepackage{mathrsfs}
\usepackage{enumerate}

\usepackage{theorem}

\newtheorem{lemma}{Lemma}     
\newtheorem{assumption}{Assumption}   
\newtheorem{theo}{Theorem}

\theoremheaderfont{\sc}\theorembodyfont{\upshape}
\theoremstyle{nonumberplain}
\theoremseparator{}
\theoremsymbol{\rule{1ex}{1ex}}

\def \E {{\bf E}}
\def \F {\mathscr{F}}
\def \ud {\mathrm{d}}
\def \co {\mathcal{O}}
\def\cX{\mathcal{X}}
\def\argmin{\hbox{argmin}}
\def\grad{\nabla}
\def\sgrad{\widetilde{\nabla}}

\def\a{\alpha}
\def\pf{{\noindent \it Proof.\ }}
\def\[{\left[}
\def\]{\right]}
\def\cO{\co}
\def\cP{\mathcal{P}}

\begin{document}

\title{Random Multi-Constraint Projection: Stochastic Gradient Methods for Convex Optimization with Many Constraints}

\author{Mengdi Wang,\ \ Yichen Chen\thanks{Mengdi Wang and Yichen~Chen are with Department of Operations Research and Financial Engineering and Department of Computer Science, Princeton University, Princeton 08544, USA.},\ \ Jialin~Liu, \ \ Yuantao~Gu\thanks{Jialin~Liu and Yuantao~Gu are with Tsinghua National Laboratory for Information Science and Technology, Tsinghua University, Beijing 100084, CHINA. 
}}

\date{\today}

\def\red#1{{\color{red}#1}}
\def\lf{\left} \def\ri{\right}
\def\qed{\hfill$\square$}
\def\CSO{$\mathcal{CSO}$}
\def\GSO{$\mathcal{GSO}$}
\def\SO{$\mathcal{SO}$}
\def\bO{O}
\def\dist{{\rm d}}
\def\smskip{\vspace{6pt}}
\def\M{M}
\def\cO{\mathcal{O}}
\def\o{\omega}

\maketitle

\begin{abstract}
Consider convex optimization problems subject to a large number of constraints. We focus on stochastic problems in which the objective takes the form of expected values and the feasible set is the intersection of a large number of convex sets. We propose a class of algorithms that perform both stochastic gradient descent and random feasibility updates simultaneously. At every iteration, the algorithms sample a number of projection points onto a randomly selected small subsets of all constraints. Three feasibility update schemes are considered:  averaging over random projected points, projecting onto the most distant  sample, projecting onto a special polyhedral set constructed based on sample points.  We prove the almost sure convergence of these algorithms, and analyze the iterates' feasibility error and optimality error, respectively. We provide new convergence rate benchmarks for stochastic first-order optimization with many constraints. The rate analysis and numerical experiments reveal that the algorithm using the polyhedral-set projection scheme is the most efficient one within known algorithms. 
\end{abstract}


\section{Introduction}
Consider the optimization problem
\begin{equation}
\label{eq-cop}
\begin{split}
&\text{minimize    }~  \Big\{ F(x) = \E[ f(x;v) ] \Big\}\\
&\text{subject to       }~ x \in \cX = \cap_{i=1}^m \cX_i,
\end{split}
\end{equation}
where $F:\Re^n\mapsto\Re$ is a continuous function, $f(\cdot;v) : \Re^n \mapsto \Re$ is a function parameterized by $v$, $\cX_i$ are closed and convex sets in $\Re^n$, $v$ is a random variable, $m$ is the total number of constraints (which is potentially a large number), and the expectation $\E [\cdot]$ is taken over the distribution of $v$. {\it Throughout this paper, we assume that $F$ is a convex function but not necessarily smooth.}

We focus on situations where $F$ is an expectation of random functions and is not available to us directly. Instead, we are given a sampling oracle that allows us to query for random realizations of the subgradients of $F$. We also focus on the situation where the feasible set is the intersection of many constraint supersets, e.g.,
$$\cX= \{ g_i(x) \leq 0,\quad \forall ~i =1,\ldots,m\} =\cap_{i=1}^m \{ g_i(x) \leq 0\},$$
where each superset is $\cX_i = \{g_i(x)\leq 0\}.$ When $m$ is a large number, it is often inefficient or even impossible to operate over $\cX$ directly. Instead, the sampling oracle is able to sample a small subset of $\{\cX_i\}_{i=1}^m$ and perform projection onto the sample constraints.

Problem (\ref{eq-cop}) is a canonical problem frequently arising from large-scale computing, stochastic optimization, machine learning, estimation and filtering.  In the case where $\cX = \Re^n$ or $\cX$ is a simple set, \emph{stochastic gradient descent} (SGD) is a popular method and has been studied extensively. SGD updates incrementally according to
\begin{equation*}
x_{k+1} = \Pi_{\cX}\lf\{ x_k - \alpha_k g(x_k,v_{k+1}) \ri\},
\end{equation*}
where $g(x_k,v_{k+1})$ is a sample realization of a subgradient of $F$ at $x_k$ and $\alpha_k$ is a positive stepsize.
SGD is one of the most important stochastic first-order methods and has drawn significant attention from various communities. 
Theoretically, it has been shown that after $k$ samples/iterations, the
average of the iterates has $\cO\lf(1/k\ri)$ optimization error for strongly convex problems, and $\cO(1/\sqrt{k})$ error for general convex problems (see Moulines and Bach \cite{moulines2011non}, Rakhlin et al. \cite{RSS11}, Shamir and Zhang \cite{ShZ12}). These convergence rates match the corresponding information-theoretic lower bounds (see e.g., Agarwal et al. \cite{ABRW12} and  Nesterov and Yudin \cite{NuY83}), suggesting that SGD is non-improvable with respect to the sample size. 
Practically, SGD is a fast algorithm that is able to process one data entry per iteration. It can be implemented as a distributed or parallel algorithm to process large-scale data sets. It can also be implemented as an online algorithm to process streaming data. The nice theoretical property and flexibility of implementation makes SGD one of the most popular methods for many machine learning applications. It has motivated extensive theoretical research as well as many new algorithms for specific applications (see \cite{ram2009incremental,GhLI,GhLII, ShZ12} for examples). 

For constrained optimization,  SGD often becomes inefficient when $\cX$ is a complicated set like $\cX = \cap_{i=1}^m \cX_i$. In particular, each iteration of SGD requires calculating the Euclidean projection onto the feasible set $\cX$, which can be computationally expensive. This is a serious limitation. Although $\cX = \cap_{i=1}^m \cX_i$ is hard to work with, it is often convenient to project a vector $x$ onto a single constraint set $\cX_i$. This has motivated the {\it random projection algorithm}, taking the form
\begin{equation}
\begin{split}
\label{alg-wmd}
x_{k+1} =& \Pi_{\cX_{\omega_k}} \{ x_k - \alpha_k g(x_k,v_{k+1}) \},
\end{split}
\end{equation}
where $\Pi_{\cX_{\omega_k}}$ denotes the projection on $\cX_{\omega_k}$ and $\omega_k$ is a random variable taking value in $\{1,\ldots,m\}.$ At each iteration, algorithm \eqref{alg-wmd} randomly picks one out of all constraint sets and finds the projection onto it. When the feasible set $\cX$ is the intersection of many simple sets (e.g., half spaces determined by linear inequalities), algorithm \eqref{alg-wmd} is able to solve large-scale problems by fast  incremental updates. The idea of incremental projection is also related to a line of works on the feasibility problem, which is to find a common point in the intersection of many convex sets (see von Neumann \cite{vNe50}, Halperin \cite{Hal62}, Gubin et al.\ \cite{GPR67}, Tseng \cite{Tse90}, Bauschke et al.\ \cite{BBL97}, Deutsch and Hundal \cite{DeH06a}, \cite{DeH06b},  \cite{DeH08}, Cegielski and Suchocka \cite{CeS08}, Lewis and Malick \cite{LeM08}, Leventhal and Lewis \cite{LeL10}, and Nedi\'c \cite{Ned10}).

Optimization algorithms using random feasibility updates were first considered by Nedi\'c \cite{Ned11}, 
and were later studied by Wang and Bertsekas in \cite{WaB12} in the context of stochastic variational inequalities. A recent paper \cite{WaB13} studied the random projection algorithm \eqref{alg-wmd} and its proximal variant, and provided a unified analytic framework for its almost sure convergence. 
There remain several open questions. First, a comprehensive convergence rate analysis addressing various situations (such as the case of strongly convex optimization) is yet to be established. Second, it is not clear how the constraint randomization scheme affects the algorithms' efficiency. Third, it would be interesting to design new algorithms that make more efficient use of the sample gradients and sample constraints.

The aim of this paper is to gain a deeper understanding of constraint randomization and to develop faster algorithms that make smart use of random constraint projections. The existing method uses one random projection per iteration, which may induce large oscillation in the iterates and cause the convergence to be slow. For this reason, we propose new algorithms that sample multiple constraint supersets per iteration. Various feasibility update schemes are considered. We are particularly interested in obtaining tight theoretical guarantees of the algorithms' performance. In addition, we pay special attention to the efficiency of various feasibility update schemes. 

The contribution of this paper are three-folded. 
\begin{enumerate}[(i)]

\item We propose a new class of {\it random multi-constraint projection algorithms} that {incrementally} process the stochastic subgradients as well as sample constraint projections. It contains the existing algorithm \eqref{alg-wmd} as a special case.
In particular, we consider three algorithms with different feasibility update steps. The first algorithm averages over multiple random projections.
The second algorithm projects the iterate onto the most distant set out of all the sample sets.
The third algorithm constructs a new polyhedral set based on the sample projected points and projects the iterates onto the polyhedral set.

\item We provide a comprehensive convergence and convergence rate analysis that addresses various assumptions and algorithm variants. We analyze the performance of these random multi-constraint projection algorithms from two aspects: \emph{convergence of feasibility error} and \emph{convergence of optimality error}. 
We provide tight estimates of the convergence rates to illustrate the efficiency of various random feasibility update schemes. A summary of our convergence rate results is given in Table \ref{table_rate}. They are the first complete convergence rate results for the class of random projection algorithms. Comparing the convergence rates in Table 1 with the non-improvable convergence rates of SGD, we discover a surprising phenomenon: {\it Constraint randomization does not slow down the convergence of stochastic first-order methods (up to constant factors).} In other words, the random projection algorithms enjoy almost the same practical advantages and theoretical guarantees as the popular SGD method.

\begin{table}[h!]
\centering
\begin{tabular}{|c|c|c|}
\hline
\hline
 & Convex Optimization  & Strongly Convex Optimization \\
\hline
Optimality Error  & $\displaystyle{\co \lf( \frac{D^2 + \lf(1+\frac1C\ri)B^2 }{\sqrt k} \ri)}$ & $\displaystyle{\co \lf(\frac{(1+\frac1{C})B^2}{\sigma^2 }\cdot \frac{\log k +1 }k\ri)}$ 
\\

\hline
Feasibility Error  &$\displaystyle{\co \lf(\frac{(1+C)B^2}{C^2} \cdot \frac{1}k\ri)}$ 
& $\displaystyle{\co \lf(\frac{(1+C)B^2}{C^2}\cdot \frac{1 }{k^2}\ri)}$ 
\\
\hline
\hline
\end{tabular}
\caption{Best known convergence rates of random multi-constraint projection algorithms. Here the constant $C$ is related to the feasibility update scheme, and the constant $B$ is a stochastic analog of the Lipschitz continuity constant of $F$.
\label{table_rate}}
\end{table}

\item  We show that by sampling multiple constraints instead of sampling a single constraint, the convergence rate is accelerated significantly, as long as one uses a good feasibility update scheme. Consider the three feasibility update schemes: the averaging scheme, the max-distance-set scheme, and the polyhedral-set scheme. We show that the max-distance-set scheme and polyhedral-set schemes converge significantly faster than the averaging scheme. Moreover, we show that the polyhedral-set algorithm outperforms the other two in a majority of practical situations (except for a worst-case situation where its performance is identical with the max-distance-set scheme), and that the sample efficiency improves as more constraint samples are used per iteration. 
Moreover, we illustrate that the efficiency of random feasibility updates strongly depends on the spatial distribution of the constraints (such as angles between the normal planes of constraint sets and the sampling distribution) as well as the initial iterate. Both analytical and experimental justifications are provided.

\end{enumerate}

The remainder of this paper is organized as follows. In Section 2, we propose the class of random multi-constraint projection algorithms with three feasibility update schemes. In Section 3, we show that the convergence  of these algorithms is an interplay of a feasibility improvement process and an optimality improvement process,  and we prove the almost sure convergence of all three algorithms. In Section 4, we study the rates of convergence of the feasibility error and the optimality error, respectively. In Section 5, we provide numerical experiments; and In Section 6, we draw the conclusions.

\paragraph{Notations} All vectors are considered as column vectors. For a vector $x\in\Re^n$, we denote by $x'$ its transpose, and denote by $\|x\| =\sqrt{x'x}$ its Euclidean norm. For a matrix $A\in\Re^{n\times n}$, we denote by $\|A\| = \max \{\|Ax\| \mid \|x\|=1 \}$ its induced Euclidean norm. For two sequences $\{a_k\},\{b_k\}$, we denote by $a_k = \bO(b_k)$ if there exists $c>0$ such that $\|a_k\| \leq c\|b_k\|$ for all $k$.
For a set $\cX\subset\Re^n$ and vector $y\in\Re^n$, we denote by 
$$\Pi_{\cX}\{ y\} = \argmin_{x\in \cX}\|y-x\|^2$$
the Euclidean projection of $y$ on $\cX$, 
where the minimization is always uniquely attained if $\cX$ is nonempty, convex and closed. For a function $f(x)$, we denote by $\grad f(x)$ its gradient at $\cX$ if $f$ is differentiable, denote by $\partial f(x)$ its subdifferential at $\cX$, and denote by $\sgrad f(x)$ some subgradient at $\cX$ (to be specified in the context). 

\section{Random Multi-Constraint Projection Algorithms}
\label{sec_algo}

In this section, we propose three random multi-projection algorithms, which are described in Algorithms 1, 2, and 3, respectively. They update using stochastic (sub)gradients and random projections onto a small subset of constraints. 
Suppose that we are given a {\it Sampling Oracle} (\SO)  such that
\begin{itemize}
\item Given $x\in\Re^n$, the \SO\ returns a random subgradient $g(x,v)$ of the objective function $F$.
\item Given $x\in\Re^n$ and integer $M>0$, the \SO\ returns $M$ random projected points $\{\Pi_{\cX_{\o_1}} (x),\ldots,\Pi_{\cX_{\o_{\M}}} (x)\}$ where the sample sets are drawn uniformly without replacement.
\end{itemize}
Here $v,\o$ are independent random variables, and $\M\ll m$ is a positive integer. Note that the \SO\ only returns the projected point $\Pi_{\cX_\o} (x)$, not the constraint set $\cX_\o$ itself. {\it We assume throughout that random variables generated by different calls to the \SO\ are independent and identically distributed.}

Our proposed algorithms update the iterates $\{x_k\}$ while interacting with the \SO. Each iteration alternates between two steps: an optimality update step (stochastic gradient descent), and a feasibility update step (random projections). The three algorithms considered in this paper use the same optimality update step, which is a straightforward stochastic gradient descent step. They differ from one another in their feasibility update steps.  See Figure 1 for graphical illustrations of the three algorithms.

\begin{algorithm}[h]\small
\caption{Random Averaging Projection Method}
\label{alg-main}
\textbf{Input:}  $x_0\in\Re^n$, \SO, integer $\M>0$, stepsizes $\{\a_k\}\subset \Re^+$.\\
\For{$ k = 0,1,2,\dots $}
{
\textbf{(1)} Sample a stochastic subgradient $g(x_k,v_{k+1})$ from the \SO\ and update by
$$
y_{k+1} = x_k - \alpha_k g(x_k,v_{k+1})\ ;
$$
\textbf{(2)} Sample $M$ projections $\{\Pi_{\cX_{\omega_{k+1,1}}}y_{k+1},\ldots,\Pi_{\cX_{\omega_{k+1,\M}}}y_{k+1}\}$ from the \SO\ and update by
$$
x_{k+1} = \frac1M \sum_{i=1}^{M}\Pi_{\cX_{\omega_{k+1,i}}}y_{k+1}\ ;
$$
}
\end{algorithm}

\begin{algorithm}[h]\small
\caption{Random Max-Set Projection Method}
\label{alg-max}
\textbf{Input:}  $x_0\in\Re^n$, \SO, integer $\M>0$, stepsizes $\{\a_k\}\subset \Re^+$.\\
\For{$ k = 0,1,2,\dots $}
{
\textbf{(1)} Sample a stochastic subgradient $g(x_k,v_{k+1})$ from the \SO\ and update by
$$
y_{k+1} = x_k - \alpha_k g(x_k,v_{k+1})\ ;
$$
\textbf{(2)} Sample $M$ projections $\{\Pi_{\cX_{\omega_{k+1,1}}}y_{k+1},\ldots,\Pi_{\cX_{\omega_{k+1,\M}}}y_{k+1}\}$ from the \SO\ and update by
$$
x_{k+1} = \Pi_{\cX_{\omega_{k+1,i^*}}}y_{k+1}\ ;$$
where 
$$
i^*=\hbox{argmax}_{i=1,\ldots,\M} \|\Pi_{\cX_{\omega_{k+1,i}}}y_{k+1}-y_{k+1}\|\ ;
$$
}
\end{algorithm}

\begin{figure}[t]
\centering
\includegraphics[width=0.45\textwidth]{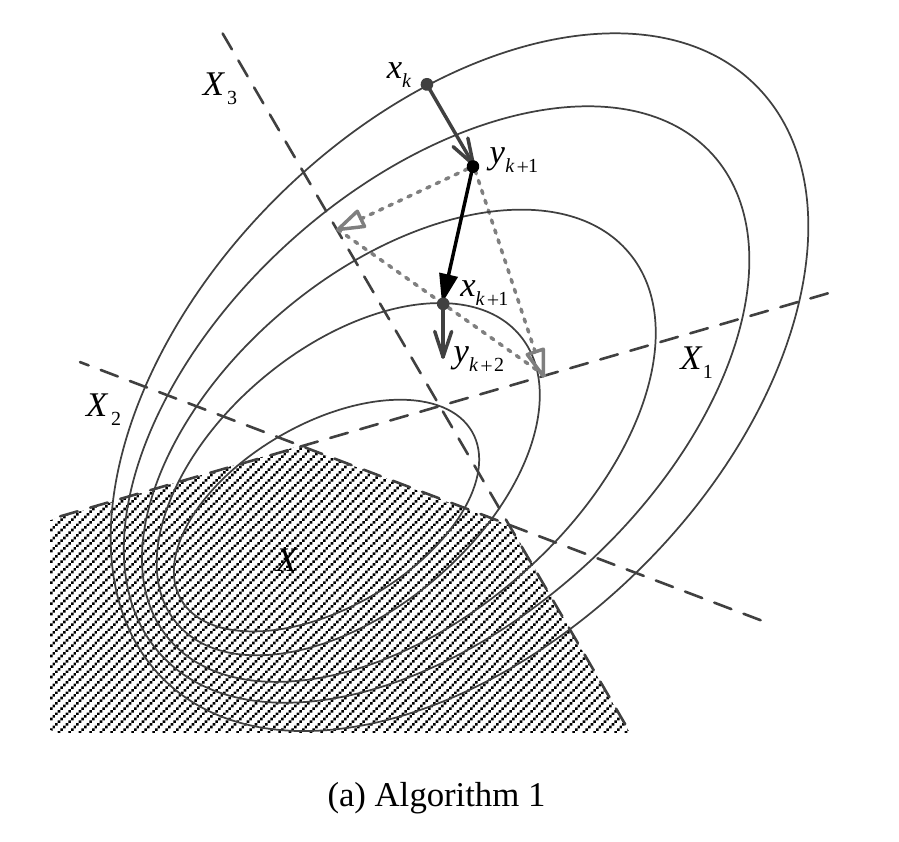}
\includegraphics[width=0.45\textwidth]{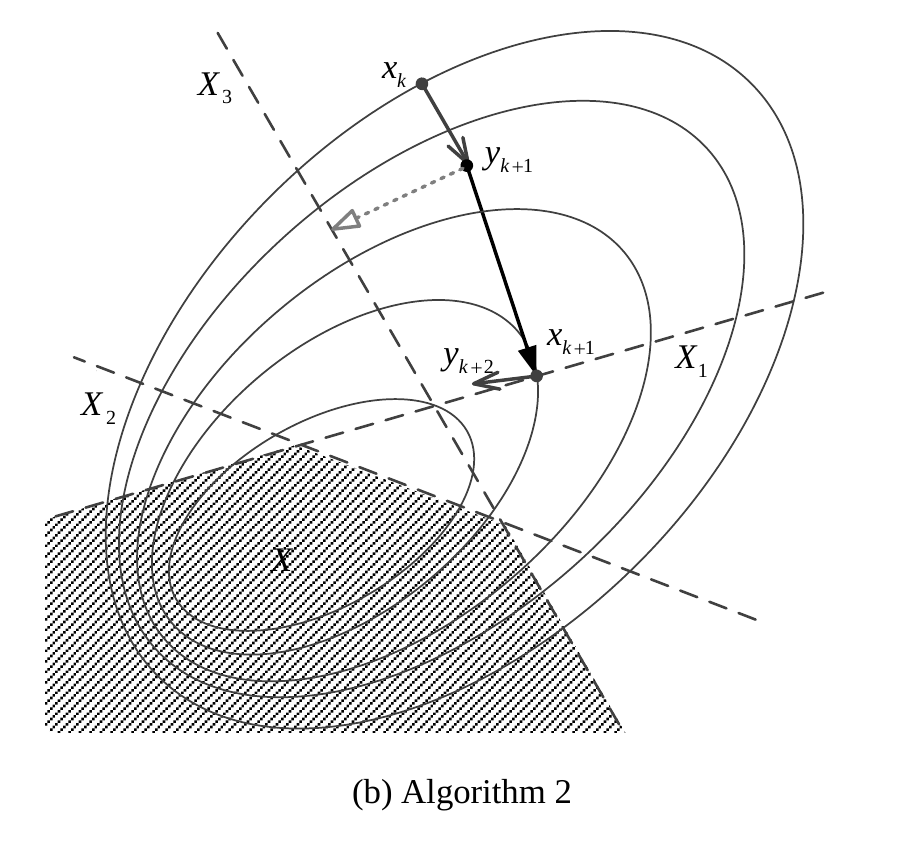}\\
\includegraphics[width=0.45\textwidth]{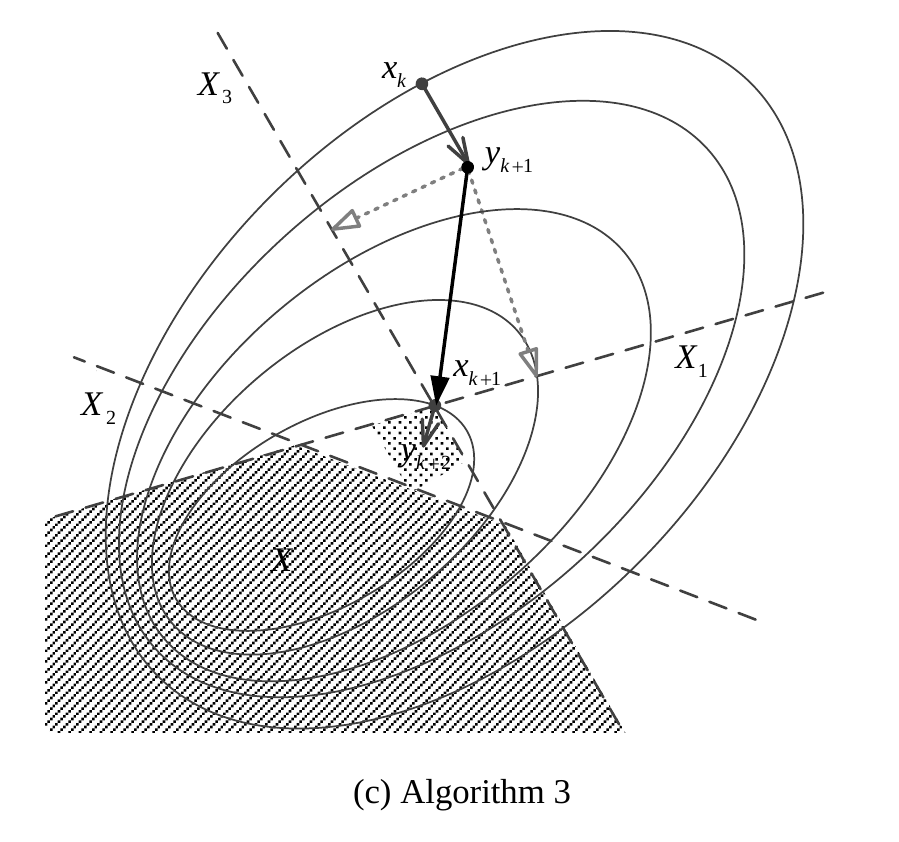}
\includegraphics[width=0.45\textwidth]{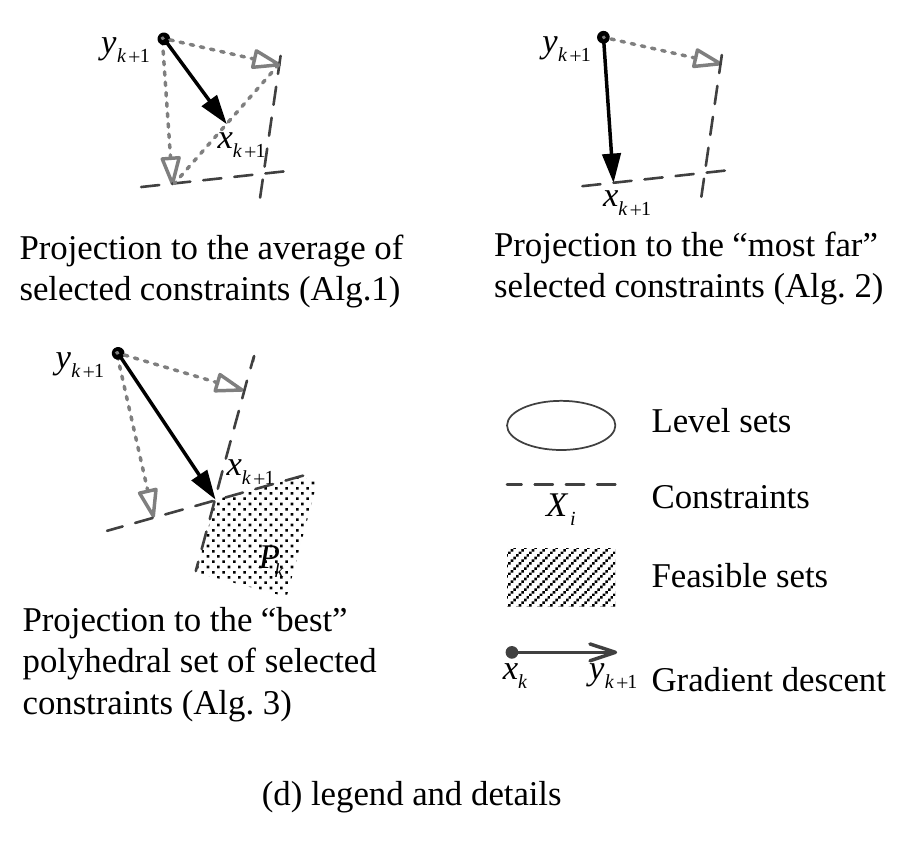}
\caption{Graphical visualization of Algorithms 1,2,3. In this example, the feasible set is $X=X_1\cap X_2\cap X_3$ and the sample constraints at the current iteration are $\{X_2,X_3\}$. }
\label{fig_icapm}
\end{figure}

Algorithm 1 takes an average of multiple random projected points. It reduces to the known algorithm \eqref{alg-wmd} when $\M = 1$.  
As illustrated by Figure \ref{fig_icapm}(a), the averaging step largely prevents the next iterate $x_{k+1}$ from randomly jumping between distant constraint sets. While improving the stability of iterates, the averaging scheme is computationally efficient, as calculating each random projection still involves only one set $\cX_i$.

Algorithm 2 chooses the most distant set out of the sample constraints. By comparing the distances between the projected points and the original point, the algorithm can easily identify the most distant set out of all samples. Then it updates by setting the next iterate to be the most distant projected point.
See Figure 1(b) for a graphical visualization of Algorithm 2.
By projecting onto the most distant set, it guarantees larger improvement towards the feasible set and faster convergence than Algorithm 1. 

\begin{algorithm}[h] \small
\caption{Random Polyhedral-Set Projection Method}
\label{alg-mul}
\textbf{Input:} $x_0\in\Re^n$, \SO, integer $\M>0$, stepsizes $\{\a_k\}\subset \Re^+$.\\
\For{$ k = 0,1,2,\dots $}
{
\textbf{(1)} Sample a random subgradient $g(x_k,v_{k+1})$ and update the iterate by
$$
y_{k+1} = x_k - \alpha_k g(x_k,v_{k+1})\ ;
$$
\textbf{(2)} Sample $M$ random projections $\{\Pi_{\cX_{\omega_{k+1,1}}}y_{k+1},\ldots,\Pi_{\cX_{\omega_{k+1,\M}}}y_{k+1}\}$ from the \SO\ and construct a system of linear inequities by letting  $$\cP_k =\{a_i' x\leq b_i,\ \  i =1,\ldots,\M\}\ ;$$
where
 $a_i = y_{k+1} - \Pi_{\cX_{k,i}} y_{k+1}$, $b_i = a_i'\Pi_{\cX_{k,i}} y_{k+1} $ for $i=1,\ldots,\M.$

\textbf{(3)} Calculate $x_{k+1}$ as 
$$x_{k+1} = \hbox{argmin}_{x\in \cP_k} \|x - y_{k+1}\|^2\ ;$$
}
\end{algorithm}

Algorithm 3 utilizes the random projection points in a different way; see Figure 1(c) for an example.
Given the random projected points and the corresponding normal hyperplanes, it constructs a new polyhedral set $\cP_k$ per each iteration. This polyhedral set $\cP_k$ is also a superset of the feasible set $\cX$, and it is a better approximation to the unknown $\cX$. Projecting onto $\cP_k$ involves minimizing a square function over a small set of linear inequalities. 
Such an extra projection step can be carried out easily, without affecting the iteration's efficiency.
In subsequent analysis, we show that the polyhedral-set projection method (Algorithm 3) is the most efficient one among all three methods.


\newpage

\section{Coupled Convergence Process}
\label{sec_rate}

In this section, we study the convergence of random multi-constraint projection algorithms. Each iteration of the algorithms alternates between a feasibility update step and an optimality update step. Accordingly, the convergence process can be decomposed into two coupled stochastic processes: convergence towards feasibility and convergence towards optimality. More specifically, we provide two recursive bounds for the optimality error and the feasibility error, respectively. We show that,  all three algorithms converge almost surely to an optimal solution of problem \eqref{eq-cop}, as long as suitable stepsizes are used.


\def\sgrad{\widetilde{\nabla}}

Throughout this paper, we make the following assumptions regarding the convex optimization problem and the sampling oracle.

\begin{assumption}\label{cond_basic}\ 
\begin{enumerate}[(a)]
\item \label{cond_basic_a}
The objective function $F$ is convex and continuous, the feasible set $\cX = \cap_{i=1}^m \cX_i$ is nonempty, closed and convex, and there exists at least one optimal solution $x^*$ to problem \eqref{eq-cop}. 
\item \label{cond_basic_c}The sample subgradients are conditionally unbiased, i.e., for any $x \in \Re^n$ and $k\geq 0$,
\begin{equation}
\E[g(x,v_0)] \in \partial F(x).
\end{equation}
\item\label{cond_basic_d} There exists a scalar $B>0$ such that
\begin{equation*}
\E[\|g(x,v_0)\|^2] \leq B^2,\qquad \forall\ x\in\Re^n.
\end{equation*}
\item There exists a scalar $D>0$ such that
$$\|x-\bar x\|\leq D,\qquad \forall\ x,\bar x\in \cX_i, \quad i =1,\ldots,m.$$
\end{enumerate}
\end{assumption}

Assumption \ref{cond_basic} is quite general. It requires that the convex optimization problem is defined over a bounded feasible set (which is conventional for convergence rate analysis) and the objective function has bounded subgradients (which implies that the objective is Lipschitz continuous). It also requires that the sample first-order information (either gradient or subgradient) is unbiased and has bounded second moments. These conditions are very mild and are satisfied in the majority of sampling applications, as long as the underlying sampling distribution has reasonable light tails. 

\begin{assumption}\label{cond_const}\ 
There exists a constant scalar $\eta\in(0,1)$ such that for all $x \in \Re^n$
\begin{equation}
\eta \|x-\Pi_{\cX}x\|^2 \leq  \max_{i=1,\dots,m}\|x-\Pi_{\cX_i}x\|^2.
\end{equation}
\end{assumption}

Assumption 2 is known as the {\it linear regularity condition}. 
It is related to crossing angles between the individual sets $\cX_i$. Intuitively speaking, it requires that the sets $\cX_i$ behave like linear sets where they intersect with one another.
This property is automatically satisfied when $\cX$ is a polyhedral set and $\cX_i$ are linear half spaces. 
It has been studied by Deutsch and Hundal \cite{DeH08} in the context of the feasibility problem, where it is assumed in order to establish linear convergence of a cyclic projection method.
The discussions in \cite{Bau96} and \cite{DeH08} mention several cases where the linear regularity condition holds. These references suggest that Assumption 2 is a mild restriction. In fact,  it is rare to find practical examples that do not satisfy this condition.

Assumption \ref{cond_const} guarantees that projecting onto a random constraint set leads to sufficient decrease of the distance to feasibility. To see this, we let $\cX_w\in\{X_1,\ldots,X_m\}$ be randomly generated by the \SO\ according to a uniform distribution. Then we have
$$\E [\|z - \Pi_{\cX_w} z\|^2] \geq \frac{\eta}m \dist^2(z),\qquad \forall\ z\in\Re^n.$$
This lowerbound will be used to establish a contractive property of the optimality error. In fact, our entire analysis applies under the following more general condition: there exists some $\rho>0$ such that 
$$\E [\|z - \Pi_{\cX_w} z\|^2] \geq \rho\cdot \dist^2(z),\qquad \forall\ z\in\Re^n.$$
This condition requires that the \SO\ finds ``representable" constraint sets ``on average." It can be satisfied even if there are an infinite number of constraints ($m=\infty$) and non-uniform sampling distributions are used.

Let us summarize some basic notations and facts before analyzing the algorithms. We denote by $\sgrad F(x)$ the particular subgradient given by
$$\sgrad F(x) = \E[g(x,v)] \in \partial F(x).$$
For a vector $x\in\Re^n$, we denote its distance to a set $Y$ by
$
\dist(x,Y) =\|x-\Pi_Y x\|,
$
and denote its distance to the feasible set $\cX$ for short by
\begin{equation*}
\dist(x) = \dist(x,X) =\|x-\Pi_{\cX}x\|.
\end{equation*}
We define $\F_k$ to be the filtration generated by random variables that are revealed up to the $k$th iteration, i.e.,
$$
\F_k = \lf \{v_t,\{\omega_{t,i}\}_{i = 1}^{\M}, x_{t},y_{t}  ~\Big\vert~ t = 1,2,\dots,k \ri\}.
$$
Note that the random variables indexed by $k$ belong to $\F_k$, such as $x_{k}$. Random variables indexed by $k+1$ belong to $\F_{k+1}$, such as the random variables $\{v_{k+1},\o_{k+1,1},\ldots\o_{k+1,\M} \}$ sampled in the $(k+1)$th iteration of the algorithm and the resulting iterates $x_{k+1},y_{k+1}$.

\subsection{Error Decomposition and Almost Sure Convergence}

In what follows, we analyze the convergence properties of Algorithms 1, 2, and 3. 
There are two types of errors associated with the iterates $\{x_k\}$: the optimality error and the feasibility error. The feasibility error is usually positive because the random projection algorithms can not guarantee $\{x_k\}$ to be feasible. Let us estimate the conditional expected optimality errors and feasibility errors, i.e.,
$$
\lf\{\E\[\|x_{k+1}-x^*\|^2 \mid \F_k\]\ri\}, \qquad \lf\{\E[\dist^2(x_{k+1}) \mid \F_k]\ri\}.  
$$
Our first main result establishes two recursive bounds for the optimality and feasibility errors, respectively.

\begin{theo}[Error Decomposition]
\label{cor_expect}
Let Assumptions \ref{cond_basic} and \ref{cond_const} hold, let $x^*$ be an arbitrary optimal solution to problem \eqref{eq-cop},
and let the sequence $\{x_k\}$ be generated by any of Algorithms \ref{alg-main}, \ref{alg-max}, or \ref{alg-mul}.  Then for all $k \geq 0$, with probability 1 that
\begin{equation}
\label{inequ_exp_10}
\begin{split}
\E[\|x_{k+1}-x^*\|^2 \mid \F_k] 
\leq & \|x_{k}-x^*\|^2+ 5B^2 \alpha_k^2  - 2\alpha_k  (F(x_k)-F^*) - \frac{C}2\dist^2(x_k) \\
\leq & \|x_{k}-x^*\|^2+ \lf(5 +\frac2C \ri)B^2\alpha_k^2  - 2\alpha_k  (F(\Pi x_k)-F^*)  ,
\end{split}
\end{equation}
and
\begin{equation}
\label{inequ_exp_2}
\begin{split}
\E[\dist^2(x_{k+1})|\F_k] \leq & \Big(1 - \frac{C}{4}\Big)\dist^2(x_k) + \Big(5+\frac{4}{C}\Big)B^2\alpha_k^2 .
\end{split}
\end{equation}
where $C = \frac{\eta}{m}$ in the case of Algorithm \ref{alg-main}, and $C = \frac{M\eta}{m}$ in the case of Algorithms \ref{alg-max} or \ref{alg-mul}. 
\end{theo}

According to Theorem 1, both feasibility error and optimality error shrink ``on average" as the algorithms proceed. The optimality error decreases at a speed determined by the stepsize $\a_k$. The feasibility error decreases according to a geometric contraction with some additive error proportional to $\a_k^2$. Indeed, the two processes $\{\|x_k-x^*\|^2\} $ and $\{\dist^2(x_{k})\}$ are entangled with each other. However, the recursive error bounds given in Theorem 1 are almost decoupled from each other. This makes it possible to study the two convergence processes separately. The proof of Theorem 1 involves tedious technical analysis, which is deferred to Section 3.2.

Now we establish the almost sure convergence of the proposed algorithms. Due to the random noise, almost sure convergence of stochastic algorithms usually require that stepsizes diminish at a rate neither too slow nor too fast. 

\begin{theo}[Almost Sure Convergence]
\label{theo_convergence}
Let Assumptions \ref{cond_basic} and \ref{cond_const} hold, let the sequence $\{x_k\}$ be generated by any of Algorithms \ref{alg-main}, \ref{alg-max}, or \ref{alg-mul}, and let the stepsize $\{\alpha_k\}$ satisfy
$$
\sum_{k=0}^{\infty}\alpha_k = \infty,\qquad \sum_{k=0}^{\infty}\alpha_k^2 < \infty.
$$
Then 
$\{x_k\}$ converges almost surely to a random point in the set of optimal solutions of problem (\ref{eq-cop}), and $\{\dist(x_k)\}$ converges almost surely to 0.
\end{theo}

The proof of Theorem 2 relies on the recursive error bounds derived in Theorem 1.
The key to the proof is a coupled supermartingale convergence argument that has been used in \cite{RoS71, WaB13}. We defer the formal proof to the next section.

\subsection{Proofs of Theorems 1 and 2}

In the rest of this section, we develop the proofs of Theorem 1 and Theorem 2 through a series of lemmas. For readers who are not concerned with the technical details, this part can be safely skipped.

\begin{lemma}[Recursive Error Bounds]
\label{lemma_iterate_basic}
Let Assumptions \ref{cond_basic} and \ref{cond_const} hold, and let $x^*$ be an arbitrary optimal solution to problem \eqref{eq-cop}.
Suppose that $\{x_k\}$ is generated by any of Algorithms \ref{alg-main}, \ref{alg-max}, or \ref{alg-mul}. Let $e_{k+1}$ be defined as
$$e_{k+1} =  \lf\{\begin{tabular}{l c}
$\frac1M\sum_{i=1}^{M}  d^2(y_{k+1}, X_{\omega_{k+1,i}})$ &  if $\{x_k\}$ is generated by Algorithm \ref{alg-main}, \\
 $\max_{i=1,\ldots,\M} d^2(y_{k+1}, X_{\omega_{k+1,i}})$ &   if $\{x_k\}$ is generated by Algorithm \ref{alg-max}, \\
 $d^2(y_{k+1}, P_{k})$ & if $\{x_k\}$ is generated by Algorithm \ref{alg-mul}.
\end{tabular}\ri.$$
\begin{enumerate}[(a)]
\item With probability 1, for all $k\geq 0$,
\begin{equation}
\label{inequ_1}
\|x_{k+1}-x^*\|^2 \leq \|x_{k}-x^*\|^2 -2\alpha_kg(x_k,v_{k+1})'(x_k-x^*) - e_{k+1} + \alpha_k^2\|g(x_k,v_{k+1})\|^2.
\end{equation}
\item With probability 1, for all $k\geq 0$ and $\epsilon>0$, 
\begin{equation}
\label{inequ_2}
\dist^2(x_{k+1}) \leq (1+\epsilon)\dist^2(x_k)+(1+1/\epsilon)\alpha_k^2\|g(x_k,v_{k+1})\|^2 -
e_{k+1}.
\end{equation}
\end{enumerate}
\end{lemma}

\pf Note that $e_{k+1}\in\F_{k+1}$.
In each case of the three algorithms, the iterate $x_{k+1}$ takes the following form
$$x_{k+1} = \sum_{i=1}^{\M} \xi_i \Pi_{Y_i} y_{k+1},$$
where $\xi\geq 0$ for all $i$ and $ \sum_{i=1}^{\M} \xi_i=1$.
In the case of Algorithm \ref{alg-main}, we have $\xi_i = 1/\M$ and $Y_i = X_{\omega_{k+1,i}}$ for all $i=1,\ldots,\M$. 
In the case of Algorithm \ref{alg-max}, we have $\xi_i = 1$ if $i = \hbox{argmax}_{i=1,\ldots,\M} \|y_{k+1} - \Pi_{\cX_{\omega_{k+1,i}}} y_{k+1}\|$ and $\xi_i = 0$ otherwise. 
In the case of Algorithm \ref{alg-mul}, we have $\xi_i = 1/\M$ such that $ Y_ i = \cP_k $ for all $i=1,\ldots,\M$. In each of the three cases, $Y_i$ is a convex set and $\cX\subset Y_i$ for all $i=1,\ldots,\M$.
Moreover, we can verify that 
\begin{equation}\label{eq-ek}
\sum_{i=1}^{\M} \xi_i \|y_{k+1} -  \Pi_{Y_i}y_{k+1}\|^2=e_{k+1},
\end{equation}
in each case of the three algorithms.

\smskip

\noindent (a) We have
$$\begin{aligned}
&\|x_{k+1}-x^*\|^2\\
 =& \lf\| \sum_{i = 1}^{\M}  \xi_i \Pi_{Y_i}y_{k+1}-x^* \ri\|^2 \\
=& \| y_{k+1}-x^* \|^2 + \lf \| \sum_{i = 1}^{\M} \xi_i \Pi_{Y_i}y_{k+1} - y_{k+1} \ri \|^2 - 2\sum_{i=1}^{\M} \xi_i (y_{k+1}-x^*)'(y_{k+1} -  \Pi_{Y_i}y_{k+1}).
\end{aligned}$$
Consider the third term on the right side. Since $x^* \in \cX \subset Y_i$, we have
$$\begin{aligned}
 (y_{k+1}-x^*)'(y_{k+1} -  \Pi_{Y_i}y_{k+1})  
 &=  (y_{k+1} -  \Pi_{Y_i}y_{k+1})'(y_{k+1} -  \Pi_{Y_i}y_{k+1})  
 +
  ( \Pi_{Y_i}y_{k+1}-x^*)'(y_{k+1} -  \Pi_{Y_i}y_{k+1})   
 \\
 &\geq 
  (y_{k+1} -  \Pi_{Y_i}y_{k+1})'(y_{k+1} -  \Pi_{Y_i}y_{k+1})   + 0
\\ & = \|y_{k+1} -  \Pi_{Y_i}y_{k+1}\|^2,
\end{aligned}$$
where the inequality uses $( \Pi_{Y_i}y_{k+1}-x^*)'(y_{k+1} -  \Pi_{Y_i}y_{k+1}) \geq 0$, which is a property of the projection $\Pi_{Y_i}$ onto the convex set $Y_i$.
It follows that
$$\begin{aligned}
\|x_{k+1}-x^*\|^2
 &\leq \| y_{k+1}-x^* \|^2 + \lf \| \sum_{i = 1}^{\M} \xi_i \Pi_{Y_i}y_{k+1} - y_{k+1} \ri\|^2 - 2\sum_{i=1}^{\M}  \xi_i \lf\|y_{k+1} -  \Pi_{Y_i}y_{k+1} \ri\|^2\\
   &\leq \| y_{k+1}-x^* \|^2+ \sum_{i=1}^{\M}  \xi_i \lf\|y_{k+1} -  \Pi_{Y_i}y_{k+1} \ri\|^2 - 2\sum_{i=1}^{\M}  \xi_i \lf\|y_{k+1} -  \Pi_{Y_i}y_{k+1} \ri\|^2\\
  &= \| y_{k+1}-x^* \|^2 - \sum_{i=1}^{\M}  \xi_i \lf\|y_{k+1} -  \Pi_{Y_i}y_{k+1} \ri\|^2\\
    &= \| y_{k+1}-x^* \|^2 - e_{k+1},
\end{aligned}$$
where the second inequality uses the Jensen inequality that 
$$
\lf \| \sum_{i = 1}^{\M} \xi_i \Pi_{Y_i}y_{k+1} - y_{k+1} \ri\|^2 
 \leq  \sum_{i=1}^{\M} \xi_i \|(\Pi_{Y_i}y_{k+1} - y_{k+1})\|^2,
$$
and the last equality uses Eq.\ \eqref{eq-ek}.
By using the definition of $y_{k+1}$ (which is identical for all three algorithms), we have
$$\| y_{k+1}-x^* \|^2 = \|x_{k}-x^* - \a_kg(x_k,v_{k+1})\|^2= \|x_{k}-x^*\|^2 -2\alpha_kg(x_k,v_{k+1})'(x_k-x^*)  + \alpha_k^2\|g(x_k,v_{k+1})\|^2.$$  
Combining the preceding relations, we obtain \eqref{inequ_1}.

\smskip

\noindent (b)
We consider the squared distance to the feasible set $\cX$. By using the property of distance function and the Jensen inequality, we obtain
\begin{equation}\label{eq-d}
\dist^2(x_{k+1}) \leq\| x_{k+1} - \Pi_{\cX} y_{k+1} \|^2
= \lf\| \sum_{i=1}^{\M}  \xi_i (\Pi_{Y_i}y_{k+1}-\Pi_{\cX} y_{k+1}) \ri \|^2
 \leq \sum_{i=1}^{\M}  \xi_i \| \Pi_{Y_i}y_{k+1}-\Pi_{\cX} y_{k+1} \|^2.
\end{equation}
For each $i$, by using the property of projection $\Pi_{Y_i}$, we have $(\Pi_{Y_i}y_{k+1}-\Pi_{\cX}y_{k+1})'(\Pi_{Y_i}y_{k+1}-y_{k+1})\leq 0$. As a result, we have
$$\begin{aligned}
\| \Pi_{Y_i}y_{k+1}-\Pi_{\cX}y_{k+1}\|^2
\leq& \|y_{k+1}-\Pi_{\cX} y_{k+1}\|^2-\|\Pi_{Y_i}y_{k+1}-y_{k+1}\|^2.
\end{aligned}$$
Then by using the basic inequality $\|a+b\|^2 = \|a\|^2+\|b\|^2 +2a'b \leq (1+1/\epsilon)\|a\|^2 + (1+\epsilon) \|b\|^2$ for any $\epsilon>0$, we obtain
$$\begin{aligned}  
\|y_{k+1}-\Pi_{\cX}y_{k+1}\|^2
\leq & \|y_{k+1}-\Pi_{\cX}x_k\|^2\\
=& \|y_{k+1}-x_k+x_k-\Pi_{\cX}x_k\|^2\\
\leq & (1+1/\epsilon)\|y_{k+1} - x_k\|^2 + (1+\epsilon)\|x_k - \Pi_{\cX}x_k\|^2\\
=&(1+1/\epsilon)\a_k^2\|g(x_k,v_{k+1})\|^2+(1+\epsilon)\dist^2(x_k).
\end{aligned}$$
where $\epsilon$ is an arbitrary scalar. Combing the preceding two relations, we obtain
$$\begin{aligned}  
\| \Pi_{Y_i}y_{k+1}-\Pi_{\cX} y_{k+1}\|^2
\leq & (1+\epsilon)\dist^2(x_k)+(1+1/\epsilon)\alpha_k^2\|g(x_k,v_{k+1})\|^2-\|\Pi_{Y_i}y_{k+1}-y_{k+1}\|^2.
\end{aligned}$$
Applying this to \eqref{eq-d}, we obtain
$$
\dist^2(x_{k+1}) \leq 
(1+\epsilon)\dist^2(x_k)+(1+1/\epsilon)\alpha_k^2\|g(x_k,v_{k+1})\|^2 - \sum_{i=1}^{\M}  \xi_i \|\Pi_{Y_i}y_{k+1}-y_{k+1}\|^2.
$$
Finally, we apply \eqref{eq-ek} to the preceding inequality and obtain \eqref{inequ_2}.
\qed

\def\Xo{X_{\omega_{k+1,i}}}

\begin{lemma} Under the assumptions of Theorem 1, there exists a constant $C\in(0,1)$ such that
$$\E\[e_{k+1} \mid \F_k \] \geq C \cdot \E\[ \dist^2(y_{k+1}) \mid \F_k\],$$
where $C = \frac{\eta}{m}$ in the case of Algorithm \ref{alg-main}, and $C = \frac{M\eta}{m}$ in the case of Algorithms \ref{alg-max} or \ref{alg-mul}. 

\end{lemma}

\pf Note the definition of $e_{k+1}$ given by Eq.\ \eqref{eq-ek}. In the case of Algorithm \ref{alg-main}, each $\Xo$ has probability $1/m$ to be the max-distance set to $y_{k+1}$ which achieves the maximum in the linear regularity condition of Assumption 2. As a result, we have 
$$\E\[e_{k+1} \mid \F_k \]  =\frac1M\sum_{i=1}^{M}  \E\[ d^2(y_{k+1}, \Xo) \mid \F_k \] \geq \frac{\eta}{m}\E\[ \dist^2(y_{k+1}) \mid \F_k\].$$
In the case of Algorithm \ref{alg-max}, the $\M$ sets are sampled according to a uniform distribution without replacement. As a result, the max-distance set within the samples has probability $\M/m$ to be the max-distance set to $y_{k+1}$ which achieves the maximum in the linear regularity condition of Assumption 2. So we have 
$$\E\[e_{k+1} \mid \F_k \]  =
\E\[\max_{i=1,\ldots,\M} d^2(y_{k+1}, \Xo) \mid \F_k\]
\geq \frac{M \eta}{m}\E\[ \dist^2(y_{k+1}) \mid \F_k\]
.$$
In the case of Algorithm \ref{alg-mul}, we have $\cP_k\subset X_{\omega_{k+1,i}}$ for all $i$, therefore 
$$\dist^2(y_{k+1}, \cP_k)\geq \max_{i=1,\ldots,\M} d^2(y_{k+1}, \Xo).$$
Thus by using the result of the second case, we have
\begin{align*}
\E\[e_{k+1} \mid \F_k \] 
& =\E\[\dist^2(y_{k+1}, \cP_k)\mid\F_k\] \\
& \geq \E\[\max_{i=1,\ldots,\M} d^2(y_{k+1}, \Xo) \mid \F_k\] \\
& \geq \frac{M \eta}{m}\E\[ \dist^2(y_{k+1}) \mid \F_k\].
\end{align*}
Note that the inequality is tight when $y_{k+1}$ violates only one constraint.
\qed

\smskip

\smskip

We are ready the develop the main proofs of Theorems 1 and 2.
\smskip

\noindent {\bf Proof of Theorem 1.}
(a) 
Applying Lemma \ref{lemma_iterate_basic} and taking conditional expectation on both sides of (\ref{inequ_1}), we have 
$$\begin{aligned} \E[\|x_{k+1}-x^*\|^2|\F_k] \leq& \|x_{k}-x^*\|^2 -2\alpha_k\E[g(x_k,v_{k+1})|\F_k]'(x_k-x^*)
\\& - \E\[ e_{k+1} |\F_k\]+ \alpha_k^2\E[\|g(x_k,v_{k+1})\|^2|\F_k].
\end{aligned}$$
According to Assumption \ref{cond_basic}(\ref{cond_basic_c}), the second term becomes
$$ 2\alpha_k\E[g(x_k,v_{k+1})|\F_k]'(x_k-x^*) = 2\alpha_k \tilde\nabla F(x_k)' (x_k-x^*)\geq 2\alpha_k (F(x_k)-F^*) ,$$
where the inequality uses the convexity of $F$ and the property of subgradients. According to Assumption \ref{cond_basic}(\ref{cond_basic_d}), the fourth term can be bounded by
$$\begin{aligned} 
\E[\|g(x_k,v_{k+1})\|^2|\F_k] 
\leq& B^2 .
\end{aligned}$$
Applying Lemma 2 and using the following fact from \cite{WaB13} Lemma 2(b):
 $$\|y-\Pi_Sy\|^2\leq 2\|x-\Pi_Sx\|^2+ 8\|x-y\|^2,\qquad \forall x,y\in\Re^n, S\subset\Re^n \hbox{ convex},$$
 we have 
\begin{equation}\label{eq-ekk}
\begin{split}
\E\[ e_{k+1} |\F_k\] 
&\geq C \E\[ \dist^2(y_{k+1})|\F_k\]  \\
&\geq  C \E\[\frac12 \dist^2(x_k) - 4\|y_{k+1}-x_k\|^2 ~\Big\vert~\F_k\] \\
&\geq \frac{C}2 \dist^2(x_k)-4\a_k^2C B^2\\
&\geq \frac{C}2 \dist^2(x_k)-4\a_k^2B^2,
\end{split}
\end{equation}
where we have used $C\leq 1$ in the last step.
Combining the preceding inequalities, we obtain for all $k \geq 0$, with probability 1 that
\begin{equation}
\label{inequ_exp_1}
\begin{split}
\E[\|x_{k+1}-x^*\|^2|\F_k] 
\leq &  \|x_{k}-x^*\|^2 + 5\alpha_k^2 B^2- 2\alpha_k (F(x_k)-F^*) -\frac{C}2 \dist^2(x_k).
\end{split}
\end{equation}
By using the convexity of $F(x)$ and the property of subgradients, we have
$$\begin{aligned}
 F(x_k) - F(x^*)
=& F (\Pi_{\cX} x_k) - F(x^*) + F(x_k) - F(\Pi_{\cX} x_k)\\
\geq & F(\Pi_{\cX} x_k) - F(x^*) + \widetilde{\nabla} F(\Pi_{\cX} x_k)'(x_k - \Pi_{\cX} x_k)\\
\geq & F(\Pi_{\cX} x_k) - F(x^*) -\| \widetilde{\nabla} F(\Pi_{\cX} x_k)\|\|x_k - \Pi_{\cX} x_k\|\\
\geq & F(\Pi_{\cX} x_k) - F(x^*)  - B \dist(x_k),
\end{aligned}$$
where the last step is based on 
$$
\| \widetilde{\nabla} F(x) \|=\|\E[g(x,v_{k+1})|\F_k]\| \leq \sqrt{\E[\|g(x,v_{k+1})\|^2|\F_k]} \leq B,\qquad \forall\ x\in\Re^n. $$
Then we obtain 
$$\begin{aligned}
\E[\|x_{k+1}-x^*\|^2|\F_k] 
\leq&  \|x_{k}-x^*\|^2 + 5\alpha_k^2 B^2 - 2\alpha_k \big(F(\Pi_{\cX} x_k) - F(x^*)\big)\\
& + 2\alpha_k  B \dist(x_k) - \frac{C}{2} \dist^2(x_k)\\
\leq&\|x_{k}-x^*\|^2 + \lf(5+ \frac2{C}\ri)B^2 \alpha_k^2 - 2\alpha_k \big(F(\Pi_{\cX} x_k) - F(x^*)\big),
\end{aligned}$$
where the second inequality uses the basic inequality
$$ 2\alpha_k B \dist(x_k) - \frac{C}{2} \dist^2(x_k) \leq \frac{2}{C} B^2 \alpha_k^2.$$

\smskip

\noindent (b)
We apply a similar analysis to Lemma \ref{lemma_iterate_basic} Eq.\ (\ref{inequ_2}). Taking total expectation on both sides of (\ref{inequ_2}) and applying \eqref{eq-ekk}, we obtain
\begin{equation*}
\begin{split}
\E[\dist^2(x_{k+1})|\F_k] \leq & (1+\epsilon)\dist^2(x_k) + (5+1/\epsilon)\alpha_k^2B^2 - \frac{C}2 \dist^2(x_k).
\end{split}
\end{equation*}
where $\epsilon$ is an arbitrary positive scalar.
Letting $\epsilon = C/4$, we have
\begin{equation*}
\E[\dist^2(x_{k+1})|\F_k] 
\leq \Big(1 - \frac{C}{4}\Big)\dist^2(x_k) + \Big(5+\frac{4}{C}\Big)\alpha_k^2 B^2.
\end{equation*}
\qed

\noindent {\bf Proof of Theorem \ref{theo_convergence}.} 
The analysis follows from Theorem 1. Now we have obtained Eq.\ \eqref{inequ_exp_10} and Eq.\ \eqref{inequ_exp_2}. By applying the \emph{Coupled Supermartingale Convergence Theorem} (\cite{WaB13} Theorem 1), we obtain that $x_k$ converges with probability 1 to a random point in the set of optimal solutions of problem \eqref{eq-cop} and $\dist(x_k)$ converges almost surely to zero.
\qed

\newpage

\section{Convergence Rate Analysis}

In this section, we analyze the rate of convergence of the random multi-constraint projection algorithms. We provide convergence rate results for both the feasibility error and the optimality error. In particular, we study the case of convex objectives and the case of strongly convex objectives separately. 

\subsection{Convergence Rate of Feasibility Error}
\label{section_rate_dis}


Let us study the feasibility error associated with the iterates generated by the algorithms. We consider the expected squared distance from the iterate to the feasible set $\cX$, i.e., 
 $$\E[\dist^2(x_k)] = \E[ \|x_k - \Pi_{\cX} x_k\|^2].$$ 
We have shown that this feasibility error decreases to zero according to a geometric contraction with an additive error. The additive error is due to the gradient descent step and dominates the convergence of feasibility error. We give a finite-time feasibility error bound in the following theorem.

\begin{theo}[Feasibility Error Bound]
\label{theo_dis}
Let Assumptions \ref{cond_basic} and \ref{cond_const} hold, let the sequence $\{x_k\}$ be generated by any of Algorithms \ref{alg-main}, \ref{alg-max}, or \ref{alg-mul}. If there exists $\bar k\geq 0$ such that $
\alpha_{k+1}^2 \geq \big(1-\frac{C}{4}\big)\alpha_k^2
$ for all $k\geq \bar k$.
Then the feasibility error satisfies
\begin{equation}
\label{theo_dis_result}\begin{split}
\E[\dist^2(x_k)] &\leq   4B^2\Big(\frac{5}{C}+\frac{4}{C^2}\Big)   \lf(2 \a_k^{2}  + \frac{C}{4}\lf(\sum^{\bar k}_{t=0} \a_t^2 \ri)  \Big(1 - \frac{C}{4}\Big)^{k-\bar k} \ri)+ \dist^2(x_0) \Big(1 - \frac{C}{4}\Big)^{k} \\
&= 8B^2\Big(\frac{5}{C}+\frac{4}{C^2}\Big) \co( \a_k^{2} ).
\end{split}
\end{equation}
where $C = \frac{\eta}{m}$ in the case of Algorithm \ref{alg-main}, and $C = \frac{M\eta}{m}$ in the case of Algorithms \ref{alg-max} or \ref{alg-mul}. 
\end{theo}

\begin{lemma}
\label{lemma_dis_induction}
Let $\{ \delta_k \}$ and $\{\alpha_k\}$ be sequences of nonnegative scalars such that
$$
\delta_{k+1} \leq (1 - \beta) \delta_k + N\alpha_k^2,\qquad \forall\ k\geq 0,
$$
where $\beta\in(0,1)$ and $N\geq 0$ are constants.
If  there exists $\bar k\geq 0$ such that $
\alpha_{k+1}^2 \geq \big(1-\frac{\beta}{2}\big)\alpha_k^2
$ for all $k \geq \bar k$, we have
$$
\delta_{k} \leq \frac{2N}{\beta}\alpha_{k}^2 + \delta_{0} (1-\beta)^k+ \lf(N\sum^{\bar k}_{t=0} \a_t^2\ri)  (1-\beta)^{k-\bar k}.
$$
\end{lemma}

\pf
For $k \geq \bar k$, we have
$$\begin{aligned}
\delta_{k+1} \leq& (1 - \beta) \delta_k + N\alpha_k^2\\
= & (1 - \beta) \delta_k + \frac{2}{\beta}\Big(1-\frac{\beta}{2}\Big)N\alpha_k^2 - \frac{2}{\beta}(1-\beta)N\alpha_k^2\\
\leq& (1 - \beta) \delta_k + \frac{2}{\beta}N\alpha_{k+1}^2 - \frac{2}{\beta}(1-\beta)N\alpha_k^2.
\end{aligned}$$
As a result, we have
$$ \delta_{k+1} - \frac{2N}{\beta}\alpha_{k+1}^2 \leq (1-\beta)\Big( \delta_{k} - \frac{2N}{\beta}\alpha_{k}^2\Big). $$
Applying the preceding relation inductively, we obtain for all $k\geq \bar k$ that
$$
\delta_{k} \leq \frac{2N}{\beta}\alpha_{k}^2 + \Big(\delta_{\bar k} - \frac{2N}{\beta}\alpha_{\bar k}^2\Big)(1-\beta)^{k-\bar k}.
$$
Moreover, we have 
$$\delta_{\bar k} \leq \delta_0 (1-\beta)^{\bar k} + N\sum^{\bar k}_{t=0} \a_t^2  $$
Combining the preceding two inequalities, we complete the proof.
\qed

\paragraph{Proof of Theorem \ref{theo_dis}.}
We follow the proof of Theorem 1. We take expectation over \eqref{inequ_exp_2} and obtain
$$ \E[\dist^2(x_{k+1})] \leq \Big(1 - \frac{C}{4}\Big)\E[\dist^2(x_k)] + \Big(5+\frac{4}{C}\Big)B^2 \alpha_k^2. $$
We apply Lemma \ref{lemma_dis_induction} to the preceding inequality and obtain (\ref{theo_dis_result}) directly.
\qed

\smskip

\paragraph{Remark.} When the stepsize $\a_k$ is a polynomial function of $1/k$ such as $\a_k=\a_0 k^{-a}$, the stepsize assumption $
\alpha_{k+1}^2 \geq \big(1-\frac{C}{4}\big)\alpha_k^2$
is satisfied for all $k$ sufficiently large. Then we may use Theorem \ref{theo_dis} and obtain
$$
\E[\dist^2(x_k)] \leq   \co\lf( \frac{B^2(1+C)}{C^2} \cdot k^{-2a}  \ri)  + \dist^2(x_0) \Big(1 - \frac{C}{4}\Big)^{k} .
$$
As $k\to\infty$, the feasibility error is dominated by the error induced by the optimality update step, so that the distance to the feasible region satisfies
$$\dist(x_k) = \cO(k^{-a}),$$ for $k$ sufficiently large, with high probability.
 
According to Theorem \ref{theo_dis}, we learn that the constant $C\in(0,1]$ plays a key role in the convergence rate of the feasibility error. A larger value of $C$ leads to faster convergence to the feasible set $\cX$. As suggested above, the constant $C$ is a useful metric that quantifies the efficiency of the feasibility update steps. It is jointly determined by the feasibility update scheme, the spatial layout of the constraints, as well as the sampling oracle. 

\subsection{Convergence Rate of Optimality Error}
\label{section_rate_1}

Now let us analyze the optimality errors associated with the iterates $\{x_k\}$.
The traditional error metric is the objective error $F(x_k) - F(x^*)$ or the expected objective error $\E[F(x_k) - F(x^*)]$.
However, due to the constraint randomization of our algorithms, the iterates are not guaranteed to be feasible. When $x_k$ is infeasible, the objective error $F(x_k) - F(x^*)$ could take negative value, so it is not a good error metric. In order to quantify the optimality error separately from the feasibility error, we will focus on the projected iterates $\{\Pi_{\cX} x_k\}.$

For simplicity of analysis, we focus on the ergodic projected iterate defined as
\begin{equation*}
\widetilde{x}_k = \frac{1}{k}\sum_{t=0}^k  \Pi_{\cX} x_t,
\end{equation*} 
and we focus on stepsizes taking the form
$\a_k = \a_0\cdot k^{-a}.$
In the next theorem, we consider the minimization of general convex objectives and provide estimates of the optimality error after $k$ iterations.

\begin{theo}[Optimality Error for Convex Objectives]
\label{theo_epo}
Let Assumptions \ref{cond_basic} and \ref{cond_const} hold, let the sequence $\{x_k\}$ be generated by any of Algorithms \ref{alg-main}, \ref{alg-max}, or \ref{alg-mul}, and let the stepsize be $\alpha_k = \a_0 k^{-\alpha}$ for some $\a_0>0$ and $a\in(0,1].$ Then for all $k\geq 0$,
\begin{equation}
\label{theo_epo_inequ_2}
\E[F(\widetilde{x}_k) - F(x^*)] \leq \left\{
\begin{aligned}
&\frac{D^2}{2\a_0}k^{a - 1} +  \lf(\frac52+\frac1C\ri)B^2 \frac{\a_0}{1-a}   k^{-a},& a\in (0,1),\\
&\frac{D^2}{2\a_0}k^{a - 1} +  \lf(\frac52+\frac1C\ri)B^2{\a_0}  {\lf(\frac{\log k+1}{k}\ri)},&\alpha=1.
\end{aligned}
\right.
\end{equation}
\end{theo}

\paragraph{Proof of Theorem \ref{theo_epo}.}
We follow the proof of Theorem 1 and take expectation on both sides of \eqref{inequ_exp_10}. We obtain
\begin{equation}
\label{proof_epo_new_2}
\E[\|x_{k+1}-x^*\|^2] 
\leq  \E[\|x_{k}-x^*\|^2] +  \lf(5+\frac2C\ri) B^2 \alpha_k^2 - 2\alpha_k \E[F(\Pi_{\cX} x_k) - F(x^*)],
\end{equation} 
Denote for simplicity that $E_k = \E[\|x_{k}-x^*\|^2]$ and $A = (5+\frac2C)B^2$. We rearrange the preceding inequality and obtain
$$2\E[F(\Pi_{\cX} x_k) - F(x^*)] \leq \frac{1}{\alpha_k}\Big(E_k - E_{k+1}\Big) + A  \alpha_k. $$
Summing the preceding inequalities from $0$ to $k$, we obtain
$$\begin{aligned}
2\sum_{t=0}^k \E[F(\Pi_{\cX} x_t) - F(x^*)] \leq& \sum_{t=0}^k \bigg(\frac{1}{\a_t}\Big(E_t - E_{t+1}\Big) + A  \a_t\bigg)\\
=& \sum_{t=1}^k\lf(\frac{1}{\a_t}-\frac{1}{\alpha_{t-1}}\ri)E_t + \frac{1}{\alpha_0}E_0 -
\frac1{\a_{k+1}} E_{k+1}+ A \sum_{t=0}^k  \a_t. 
\end{aligned}$$
Since $\E[\|x_k - x^*\|^2]\leq D^2$ for all $k$ (from Assumption 1(d)), we obtain 
$$\begin{aligned}
2\sum_{t=0}^k \E[F(\Pi_{\cX} x_t) - F(x^*)] 
\leq & \sum_{t=1}^k\big(\frac{1}{\a_t}-\frac{1}{\alpha_{t-1}}\big)D^2 + \frac{1}{\alpha_0}D^2 + A \sum_{t=0}^k \a_t.\\
=& \frac{D^2}{\alpha_k} + A \sum_{t=0}^k\a_t  .
\end{aligned}$$
Letting $\alpha_k = \a_0 k^{-a}$, we have
$$\begin{aligned}
\sum_{t=0}^k t^{-a}\leq 1 + \int_{x=1}^{k}x^{- a}\ud x \leq \left\{
\begin{aligned}
&\frac{1}{1-a}k^{1-a},& a \in (0,1),\\
&1+\log{k},& a = 1.
\end{aligned}
\right.
\end{aligned}$$
Then we have
$$
\frac{1}{k} \sum_{t=0}^k \E[F(\Pi_{\cX} x_t) - F(x^*)] 
\leq \left\{
\begin{aligned}
&\frac{D^2}{2\a_0}k^{a - 1} +  \frac{\a_0}{2(1-a)} A  {\big(k^{-a}\big)},& a\in (0,1),\\
&\frac{D^2}{2\a_0}k^{a - 1} + \frac{\a_0}2 A  {\lf(\frac{\log k+1}{k}\ri)},&\alpha=1.
\end{aligned}
\right.
$$
Finally, by using the convexity of $F$ and the Jensen inequality, we obtain
$ \E[F(\frac{1}{k} \sum_{t=0}^k \Pi_{\cX} x_t) - F(x^*)]  \leq \frac{1}{k} \sum_{t=0}^k \E[F(\Pi_{\cX} x_t) - F(x^*)] 
$ and complete the proof.
\qed

\smskip

\paragraph{Remark.} In order to  minimize the error bound, the best stepsize choice is 
$$\a_k ={\a_0}\cdot\frac1{\sqrt k}.$$
The corresponding optimality error bound becomes
$$\E[F(\widetilde{x}_k) - F(x^*)] \leq \co\lf( \frac{(1+\frac1C)B^2 + D^2 }{\sqrt{k}}
\ri).$$
Note that when $\a_k$ are chosen in this way, the stepsize condition $\sum^{\infty}_{k=0}\a_k <\infty$ required by Theorem 2 does not hold. In other words, when the stepsize are optimized for the expected optimization error, the iterates happen to {\it not} converge almost surely. 



%
%
%
%

\smskip

Next we consider the case of strongly convex objectives. We say a function $f$ is $\sigma$-strongly convex if 
there exists a constant $\sigma > 0$ such that
$$f(x) \geq f(y) + \sgrad f(y)' (x-y) + \frac{\sigma^2}2 \|x-y\|^2,\qquad \forall\ x,y\in\Re^n,$$
where $\sgrad f(y)$ is an arbitrary subgradient of $f$ at $y$.
As a result, we also have 
\begin{equation*}
\label{equ_strong_monotone}
(x-y)'(\sgrad f(x) - \sgrad f(y)) \geq \sigma \|x-y\|^2,\qquad \forall~x,y\in\Re^n.
\end{equation*}
When the objective function $F$ is strongly convex, there exists a unique solution $x^*$ to the constrained optimization problem. As a result, we may use the expected squared distance to $x^*$ as a metric of optimality error. In the next theorem, we analyze the convergence rate in terms of the expected squared solution error $\E[\|x_k - x^*\|^2]$.

\begin{theo}[Optimality Error for Strongly Convex Objectives]
\label{theo_rate_2} Suppose that $F$ is $\sigma$-strongly convex. Let Assumptions \ref{cond_basic} and \ref{cond_const} hold, let the sequence $\{x_k\}$ be generated by any of Algorithms \ref{alg-main}, \ref{alg-max}, or \ref{alg-mul}. Let the stepsize be $$\alpha_k = \frac1{2\sigma (k+1)},$$
then
\begin{equation}
\label{theo_rate_2_inequ}
\E[\|x_k - x^*\|^2]\leq \frac{(5+\frac2{C})B^2}{4\sigma^2 }\cdot\frac{1+\log (k)}{k}.
\end{equation}
\end{theo}

The proof of Theorem 5 uses the following lemma. We recall it for clarity.

\begin{lemma}[Lemma 2.1 of \cite{nedic2001convergence}]
\label{lemma_error_induction}
Let $\{ \delta_k \}$ be a sequence of nonnegative scalars such that for all $k \geq 0$
$$ \delta_{k+1} \leq (1-\frac{p}{k+1}) \delta_k + \frac{d}{(k+1)^2}, $$
for some $p>0$ and $d>0$. Then 
$$
\delta_k \leq \lf\{
\begin{tabular}{l c}
$\frac1{(k+1)^p} \lf(\delta_0 + \frac{2^pd(2-p)}{1-p}\ri)$ & if $p<1$,\\
$\frac{d(\log(k)+1)}{k}$ & if $p=1$,\\
$\frac1{(p-1){(k+1)}}\lf(d+\frac{(p-1)\delta_0-d}{(k+1)^{p-1}}\ri) $ & if $p>1$.\\
\end{tabular}
\ri.$$
\end{lemma}

\noindent {\bf Proof of Theorem \ref{theo_rate_2}.}
We recall Eq.\ (\ref{inequ_1}) which follows from Lemma 1:
$$
\|x_{k+1}-x^*\|^2 \leq \|x_{k}-x^*\|^2 -2\alpha_kg(x_k,v_{k+1})'(x_k-x^*) - e_{k+1} + \alpha_k^2\|g(x_k,v_{k+1})\|^2.
$$
Let $h_k = g(x_k,v_{k+1}) - \sgrad F(x_k)$. We analyze the second term and obtain
$$\begin{aligned} 
g(x_k,v_{k+1})'(x_k - x^*) =& \widetilde{\nabla} F(x_k)'(x_k - x^*) + (g(x_k,v_{k+1}) - \widetilde{\nabla} F(x_k))'(x_k - x^*) \\
=& \widetilde{\nabla} F(x_k)'(x_k - x^*) + h'_k(x_k - x^*)\\
=& (\widetilde{\nabla} F(x_k) - \widetilde{\nabla} F(x^*))'(x_k - x^*) + \widetilde{\nabla} F(x^*)'(x_k - x^*) + h'_k(x_k - x^*),
\end{aligned}$$
where the first term can be bounded using the strongly convexity assumption as follows
$$(\widetilde{\nabla} F(x_k) - \widetilde{\nabla} F(x^*))'(x_k - x^*) \geq \sigma \|x_k - x^*\|^2,$$
and the second term can be bounded using the optimality of $x^*$ and Assumption \ref{cond_basic}(\ref{cond_basic_d}) as follows
$$\begin{aligned}
\widetilde{\nabla} F(x^*)'(x_k - x^*) 
\geq& \widetilde{\nabla} F(x^*)'(x_k - \Pi_{\cX} x_k)\\ 
\geq& -\|\widetilde{\nabla} F(x^*)\| \|x_k - \Pi_{\cX} x_k\|\\
\geq& - \E[\|g(x^*,v_{k+1})\||\F_k]\dist(x_k) \\
\geq& -B \dist(x_k),
\end{aligned}$$ 
where the first inequality uses the fact $\sgrad F(x^*)'(x-x^*) \geq 0$ for all $x\in \cX$.
We combine the preceding relations and obtain
\begin{equation}
\label{inequ_3}
\begin{split}
\|x_{k+1}-x^*\|^2 
\leq& (1-2\alpha_k\sigma)\|x_{k}-x^*\|^2 - 2\alpha_k h'_k (x_k - x^*) + 2B\alpha_k \dist(x_k) \\
&- e_{k+1} + \alpha_k^2\|g(x_k,v_{k+1})\|^2.
\end{split}
\end{equation} 
Taking conditional expectation on both sides of (\ref{inequ_3}), and using $\E[h_k|\F_k] = 0$ and \eqref{eq-ekk}, we obtain
$$\E[\|x_{k+1}-x^*\|^2|\F_k] \leq (1-2\alpha_k\sigma) \|x_{k}-x^*\|^2 + 2B\alpha_k \dist(x_k) - \frac{C}{2}\dist^2(x_k) + 5  B^2\alpha_k^2.$$
Taking expectation on both sides and using the fact
$$2B\alpha_k \dist(x_k) - \frac{C}{2}\dist^2(x_k) \leq \frac{2B^2}{C}\alpha_k^2,$$
we have
\begin{equation}
\E[\|x_{k+1}-x^*\|^2] \leq (1-2\alpha_k\sigma) \E[\|x_{k}-x^*\|^2] + \Big(5 B^2+\frac{2B^2}{C}\Big)\alpha_k^2.
\end{equation}
By applying Lemma \ref{lemma_error_induction} to the preceding inequality, we obtain the error bound.
\qed


\subsection{Summary and Discussions}

We summarize the convergence rates results in Table 1, which is recalled here for convenience. In this table, the convergence rates are obtained using specific stepsizes such that the optimality error is minimized. 

\begin{table*}[h!]
\centering
\begin{tabular}{|c|c|c|}
\hline
\hline
 & Convex Optimization  & Strongly Convex Optimization \\
\hline
Optimality Error  & $\displaystyle{\co \lf( \frac{D^2 + \lf(1+\frac1C\ri)B^2 }{\sqrt k} \ri)}$ & $\displaystyle{\co \lf(\frac{(1+\frac1{C})B^2}{\sigma^2 }\cdot \frac{\log k +1 }k\ri)}$ 
\\

\hline
Feasibility Error  &$\displaystyle{\co \lf(\frac{(1+C)B^2}{C^2} \cdot \frac{1}k\ri)}$ 
& $\displaystyle{\co \lf(\frac{(1+C)B^2}{C^2}\cdot \frac{1 }{k^2}\ri)}$ 
\\
\hline
\hline
\end{tabular}
\end{table*}
\noindent In this table, the constant $B$ is a stochastic analog of the Lipschitz continuity constant of $F$, and it also captures the variance in the sample gradients. The constant $C$ is related to the feasibility update scheme. It is determined jointly by the sampling distribution and the spatial layout of the constraints.

Now let us compare Algorithms 1,2, and 3. The convergence rates of the three algorithm differ only in $C$. We have shown that the constant $C$ associated with Algorithms 2 and 3 are generally much larger than that of Algorithm 1. It implies that Algorithms 2 and 3 exhibit faster convergence than Algorithm 1. We provide numerical validation for this in Section 5.

In what follows, we focus on Algorithm 3. We have shown that the worst-case performance of Algorithm 3 is the same with Algorithm 2. This happens for example when there exists only one violated constraint. In this case, projecting onto the most distant set is identical with projecting onto the polyhedral constructed from sampled constraints. However, this situation occurs rarely, meaning that our convergence rate for Algorithm 3 is very conservative. 

In fact, we argue that Algorithm 3 generally achieves significantly faster rate of convergence as long as there are more than one violated constraints with high probability.
To see this, we provide a refined argument to improve the lowerbound of $e_{k+1}$ given by Lemma 2 as follows. By using the optimality of projection, we have
$$e_{k+1} = \dist^2(y_{k+1}, \cP_k ) \geq \frac{1}{\| A_k'A_k\|} \sum_{i\in I_k}\dist^2(y_{k+1}, X_{\o_{k+1,i}}) ,$$
where each row of $A_k$ is the unit normal vector of an active sample constraint at the projected point $\Pi_{\cP_k}y_{k+1}$, and $I_k$ is the index set of active sample constraint at the projection with $|I_k|=\hbox{row}(A_k)$.
When all the active constraints are equally far away from $y_{k+1}$, we further have
$$e_{k+1} \geq \frac{\hbox{row}(A_k)}{\| A_k'A_k\|} \max_{i=1,\ldots,M}\dist^2(y_{k+1}, X_{\o_{k+1,i}}) .$$
By using the convexity of $\|\cdot\|$, we can see that the improvement factor
$$ \theta(y_{k+1},\cP_k) := \frac{\hbox{row}(A_k)}{\| A_k'A_k\|} \geq 
 \frac{\hbox{row}(A_k)}{\sum^{\hbox{row}(A_k)}_{i=1}\| A_k(i,\cdot)'A_k(i,\cdot)\|} =1.$$
The improvement factor $\theta(y_{k+1},\cP_k)  $ is related to the location of $y_{k+1}$ as well as the spatial distribution of $\cP_k$. In the case when $\cP_k$ is the intersection of two or more nearly-parallel half spaces (see Figure 1(c) for an example), we may have $\theta(y_{k+1},\cP_k) \gg 1 $, achieving huge benefit by projecting onto $\cP_k$ instead of the max-distance set.

\begin{figure}\centering\includegraphics[width=0.5\textwidth]{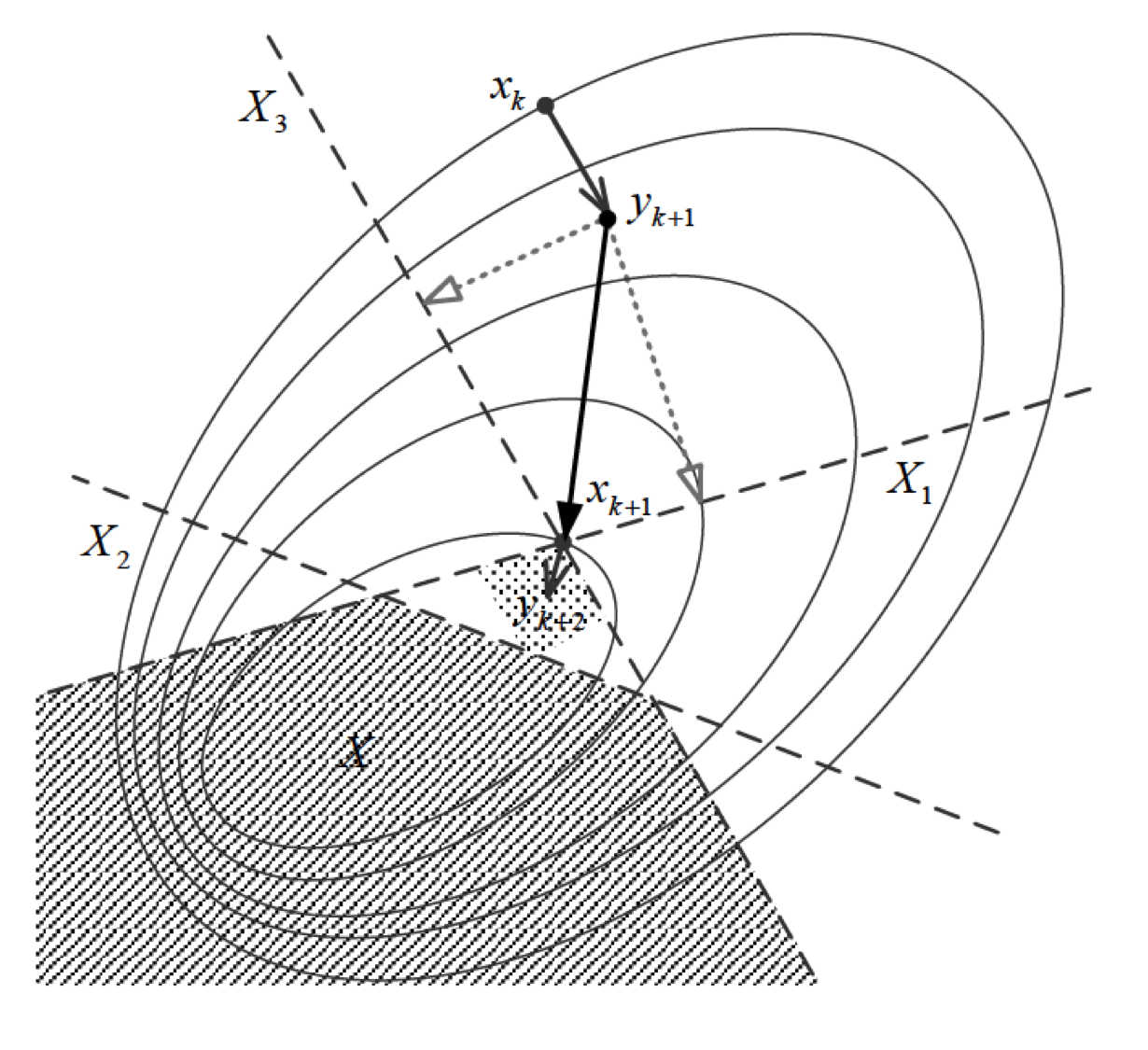}
\caption{Projecting onto the polyhedral set $\cP_k= X_1\cap X_3$ reduces the feasibility error much more than projecting onto the max-distance set $X_3$. The improvement becomes more significant when the crossing angle between $X_1$ and $X_3$ becomes smaller.}
\end{figure}
We remark that the error bounds proved in this section are tight. It would be cumbersome to incorporate the refined analysis into the rate estimates. We emphasize that, Algorithm 3 converges much faster in practice, as long as there are many constraints scattered in the space and the initial iterate violates many of them. To conclude, we have the following preferences
$$\hbox{Polyhedral Projection (Alg.\ 3)}\ \succeq\ \hbox{Max-Distance Projection (Alg.\ 2)}\ \succ\ \hbox{Average Projection (Alg.\ 1)},$$
in terms of the algorithm's convergence rate and sample complexity. Our numerical experiments further validate this result.

\smskip

\section{Numerical Results}
    In this section, we conduct two numerical experiments to justify our algorithms. In the first experiment, we test the algorithms on problems with sphere-like constraints. The results show that the max-distance-set scheme (Algorithm 2) and the polyhedral-set scheme (Algorithm 3) largely outperform the averaging scheme (Algorithm 1) and the baseline algorithm \eqref{alg-wmd}. They also show that the polyhedral-set scheme performs significantly better than all other schemes when the constraints have an ``irregular" spatial distribution.      
   In the second experiment, we apply the algorithms to a support vector machine problem. The results match very well with the error bounds predicted by theorems in Section 4. Throughout the experiments, the polyhedral-set scheme (Algorithm 3) demonstrates the best convergence properties in terms of optimization error, feasibility error, iteration efficiency, as well as sample efficiency.

\subsection{Experiments on Sphere-Like Constraints}

\begin{figure}[htb]
    \centering
    \begin{tabular}{ccc}
        {\includegraphics[width=0.33\textwidth]{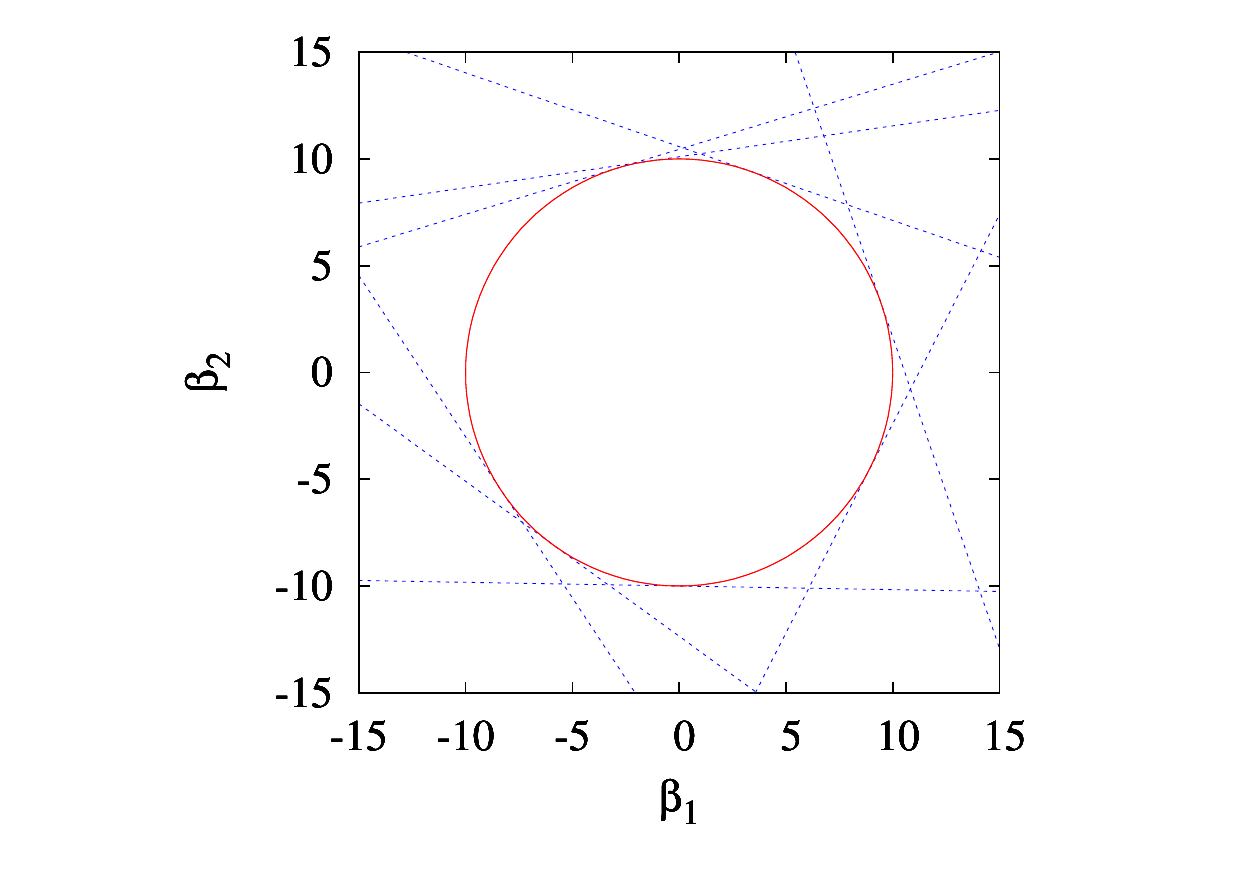}}&
        {\includegraphics[width=0.33\textwidth]{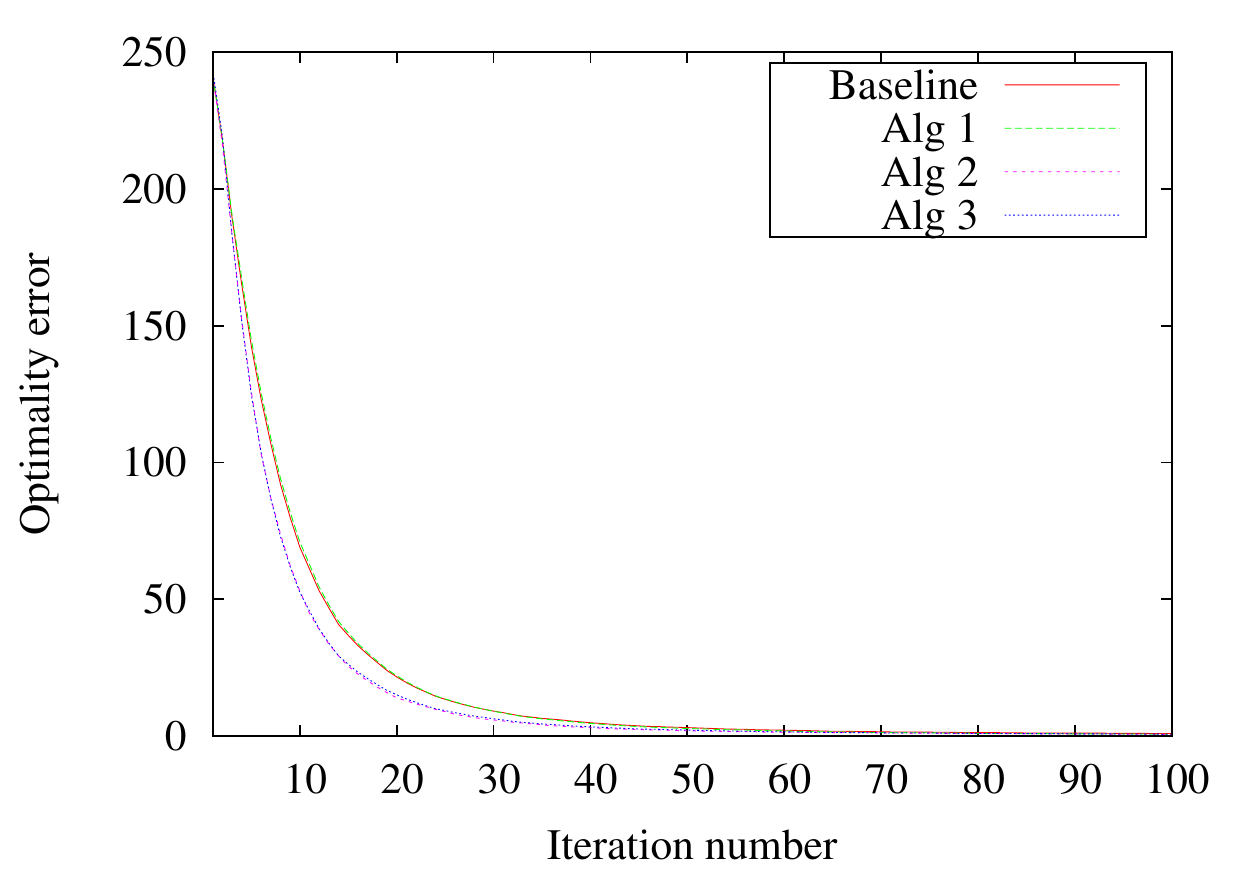}}&
        {\includegraphics[width=0.33\textwidth]{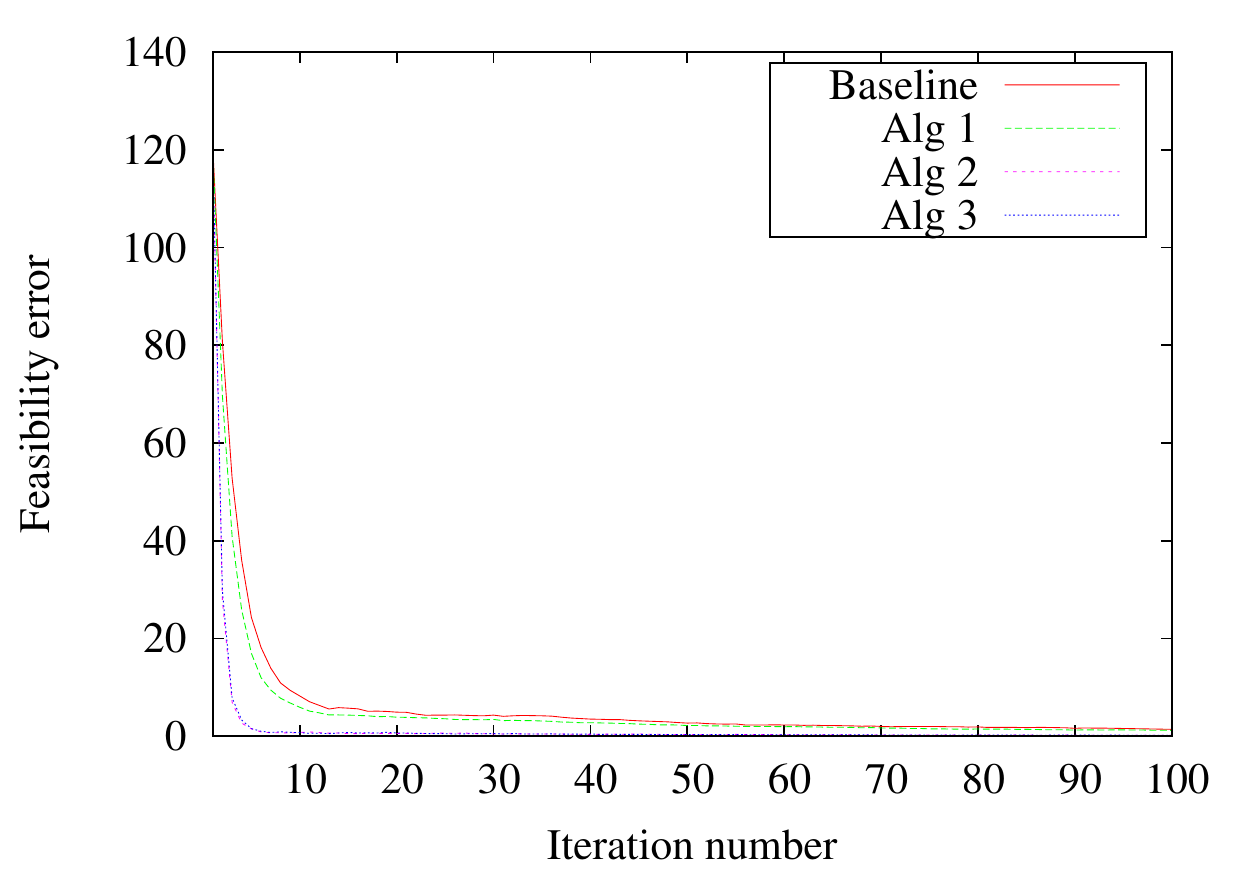}}\\ 
    \end{tabular}
    \caption{Convergence of random multi-constraint projection algorithms in experiment (i) ($d=2, M=5$ and $m=300$.) Left: Illustration of the sphere constraint. Middle: Mean optimality error. Right: Mean feasibility error. }
    \label{fig1}
\end{figure}

\begin{figure}[htb]
    \centering
    \begin{tabular}{ccc}
        {\includegraphics[width=0.33\textwidth]{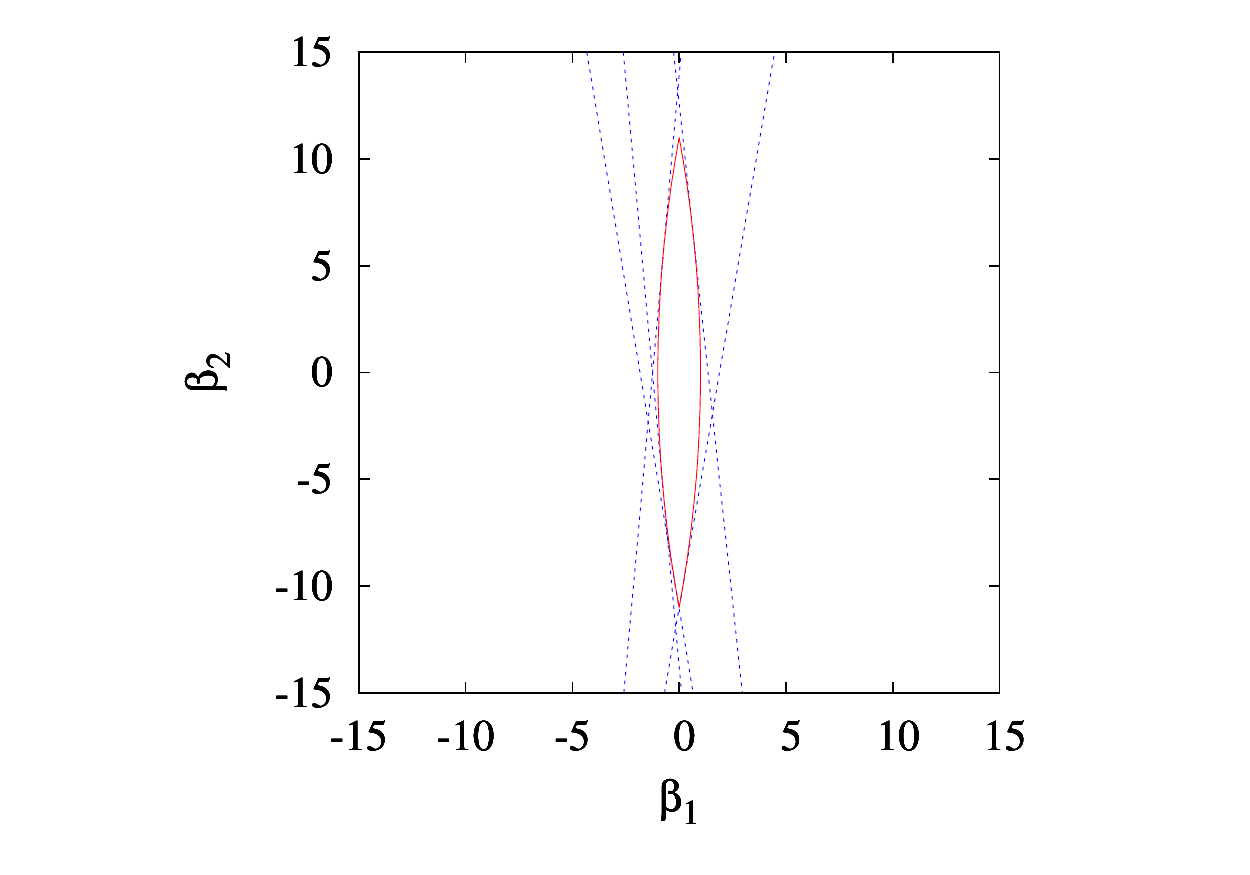}}&
        {\includegraphics[width=0.33\textwidth]{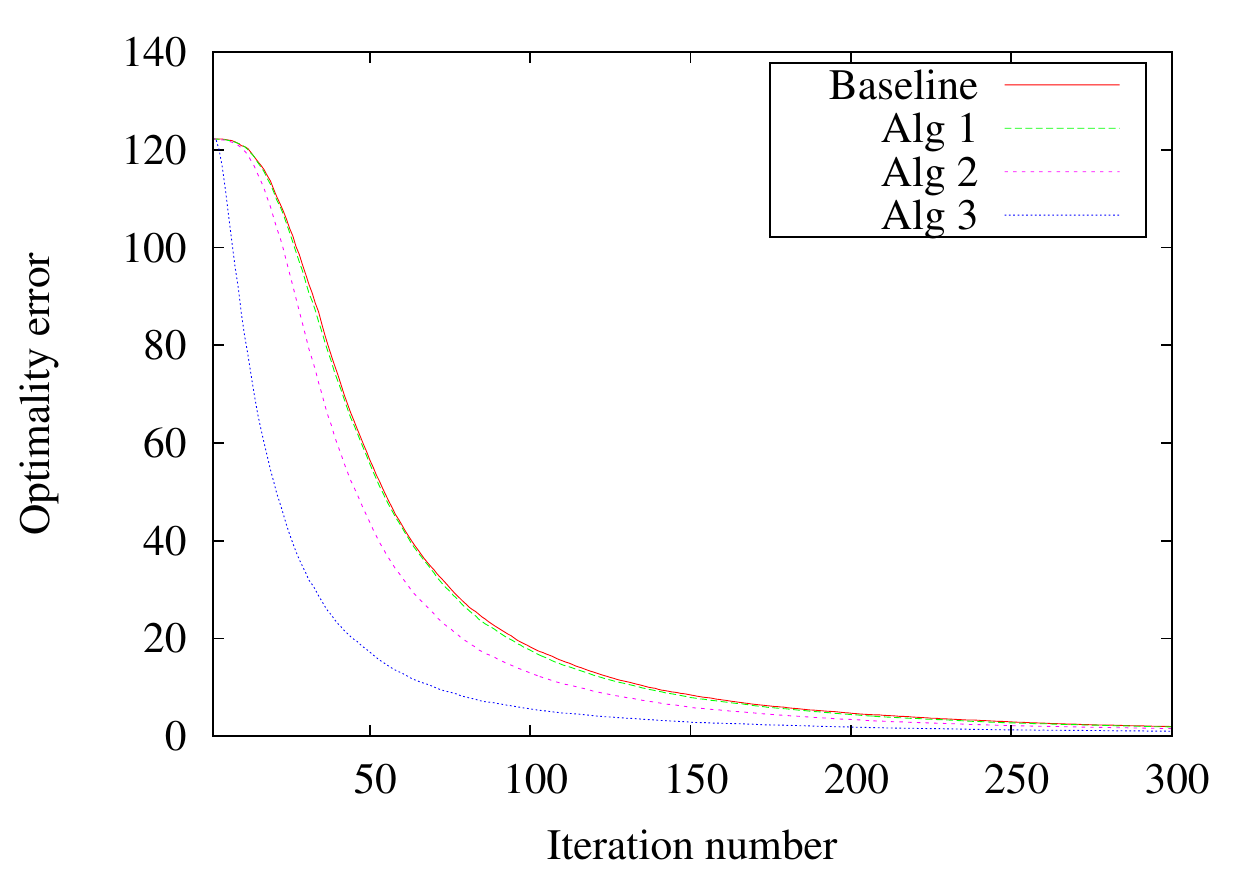}}&
        {\includegraphics[width=0.33\textwidth]{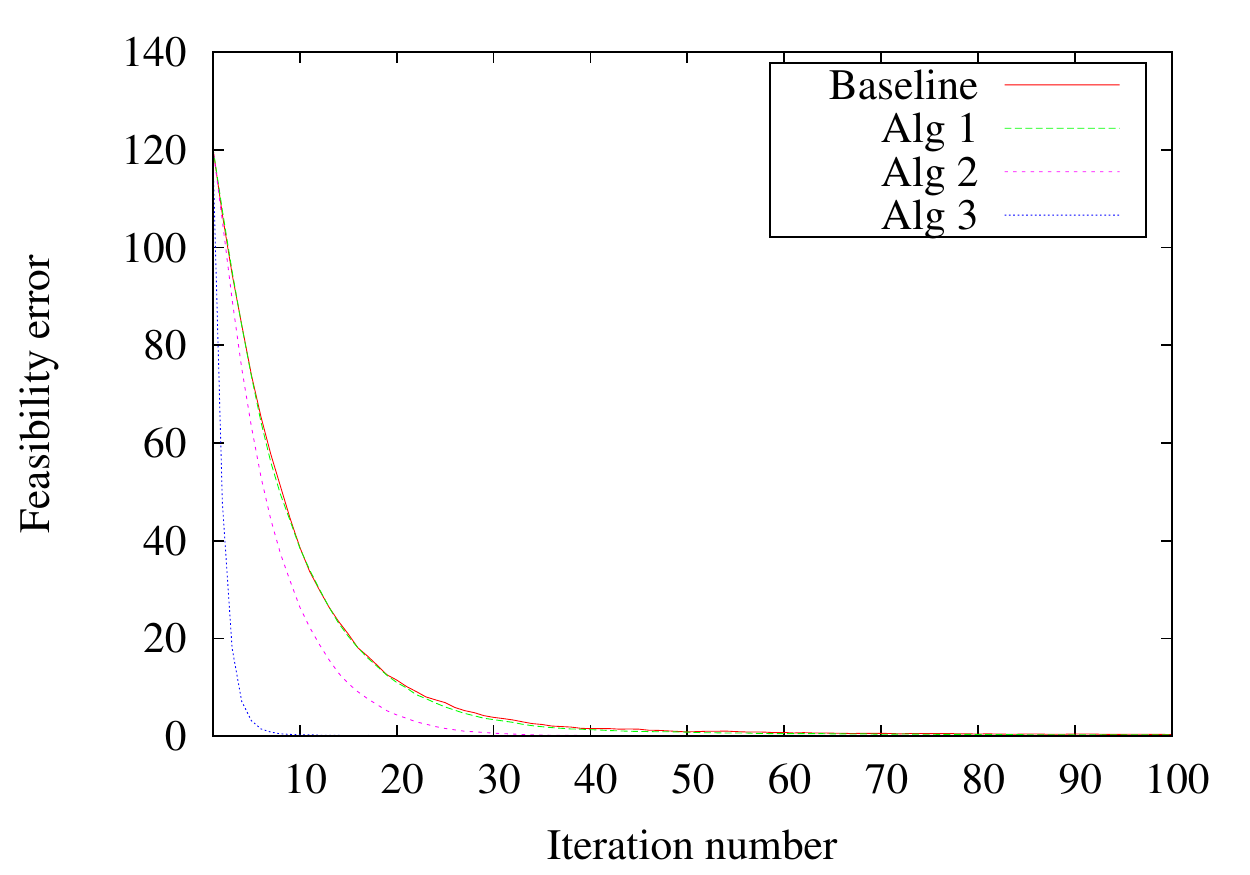}}\\ 
    \end{tabular}
    \caption{Convergence of random multi-constraint projection algorithms in experiment (ii) ($d=2, M=5$ and $m=300$.) Left: The intersection of two spheres. Middle: Mean optimality error. Right: Mean feasibility error. }
    \label{fig:fig2}
\end{figure}

We consider the following online regression problem
\begin{equation*}
    \begin{array}{ll}
        \min &\textbf{E} \left[\|Y- X^T \beta\|^2\right] \\
        \text{s.t.} & \beta\in\cX, 
    \end{array}
    \label{exp:fea:equ1}
\end{equation*}
where $X\in \mathbb{R}^d$ is a Gaussian random variable with distribution $N(0,I_d)$ and $Y=X^T \beta^*+\eta$ with $\eta\sim N(0,10)$ and some randomly generated $\beta^*$. 
We consider two different sets of constraints $\cX$: 
\begin{itemize}
\item[(i)] A system of linear constraints given by cutting planes of a sphere; as illustrated in Figure \ref{fig1}. The feasible set $\cX$ is approximately a sphere centered at 0. 
\item[(ii)]  A system of linear constraints given by cutting planes of two spheres that nearly ``touch" each other; as illustrated in Figure \ref{fig:fig2}. The feasible set $\cX$ is approximately the intersection of two spheres with radius $61$, centered at $(-60,0)$ and $(60,0)$ respectively. 
\end{itemize}
The constraints in (i) are ``regular" in the sense that constraint supersets (i.e., linear halfspaces) cross one another at large angles, corresponding to a large constant in the linear regularity condition (Assumption 2) and a large feasibility improvement constant $C$ (Lemma 2). In contrast, the constraints in (ii) have an ``irregular" distribution in the sense that some constraints cross one another at very small angles, corresponding to a small constant in the linear regularity condition and a small value of $C$.

We apply the proposed stochastic algorithms to recover the optimal solution $\beta_\text{OPT}$ from random samples of $(X_i,Y_i)$ and sample constraint cutting planes. 
We test Algorithm 1, 2, 3 with $M=5$ and the baseline algorithm \eqref{alg-wmd} with $M=1$ ($M$ is the number of sample constraints used per iteration.) 
We let the stepsize be $1/(k+10)$ and test each algorithm and parameter setting for $500$ trial runs.
Figures \ref{fig1} and \ref{fig:fig2} plot the trajectories of the mean optimality error $\|\beta_t-\beta_\text{OPT}\|_2^2$ and the mean feasibility error $\|\beta_t-\Pi_{\cX}\beta_t\|_2^2$, for experiments (i) and (ii) respectively.


In experiment (i), Algorithms 2 and 3 demonstrate similar performances  in terms of both the optimality error and the feasibility error, outperforming Algorithm 1 and the baseline. This is consistent with our theory given in Section 4.
The reason why Algorithms 2 and 3 perform similarly is due to the sphere constraint. When using the polyhedral-set projection scheme, only one cutting plane out of many samples would be active after the projection, making it identical with the max-distance-set projection scheme. As a result, Algorithm 2 and Algorithm 3 exhibit similar performances on the sphere constraint. This is an example where our worst-case error bounds are tight.

In experiment (ii), Algorithm 3 demonstrates a significantly better performance than the rest of the algorithms. This is because the constraint sets are more irregular. Consider the situation where the sample constraints contain one cutting plane from the left sphere and one from the right sphere. According to the analysis of Section 4.3, the polyhedral-set algorithm enjoys a large improvement factor, as long as the two sample cutting planes intersect at a sharp angle. This is an example where Algorithm 3 substantially outperforms the worst-case error bounds. 


\subsection{Experiments on Support Vector Machine}

The support vector machine problem is to find the best linear hyperplane that separates labeled training data $(x_i,y_i)_{i=1}^m$ into two classes. When the data are separable, it can be formulated as the minimization problem
\begin{equation*}\begin{split}
        \min_{\beta\in\Re^d}\ \  &\frac{1}{2}\|\beta\|^2\\
    \text{s.t. }\ & y_i \cdot x_i' \beta\geq 1,\\
    &  i =1,\ldots,m.
    \end{split}
    \label{exp2:equ1}
\end{equation*}
In our experiment, we generate $m$ pairs of data points  $(x_1,y_1),\dots,(x_m,y_m)$, with $x_i\in \mathbb{R}^d$ and $y_i \in \{-1,1\}$. The data points $\{x_i\}_{i=1}^m$\ are generated randomly from a mixture of Gaussian distribution. The labels $\{y_i\}_{i=1}^m$ are generated such that the data are separable and the optimization problem is feasible. 
In the support vector machine problem, each constraint naturally corresponds to a single data point. The proposed stochastic algorithms with random feasibility updates can be viewed as online training methods for the classification problem. They are able to solve the problem by processing the data points one by one.

\begin{figure}[htb]
    \centering
    \begin{tabular}{ccc}
        {\includegraphics[width=0.33\textwidth]{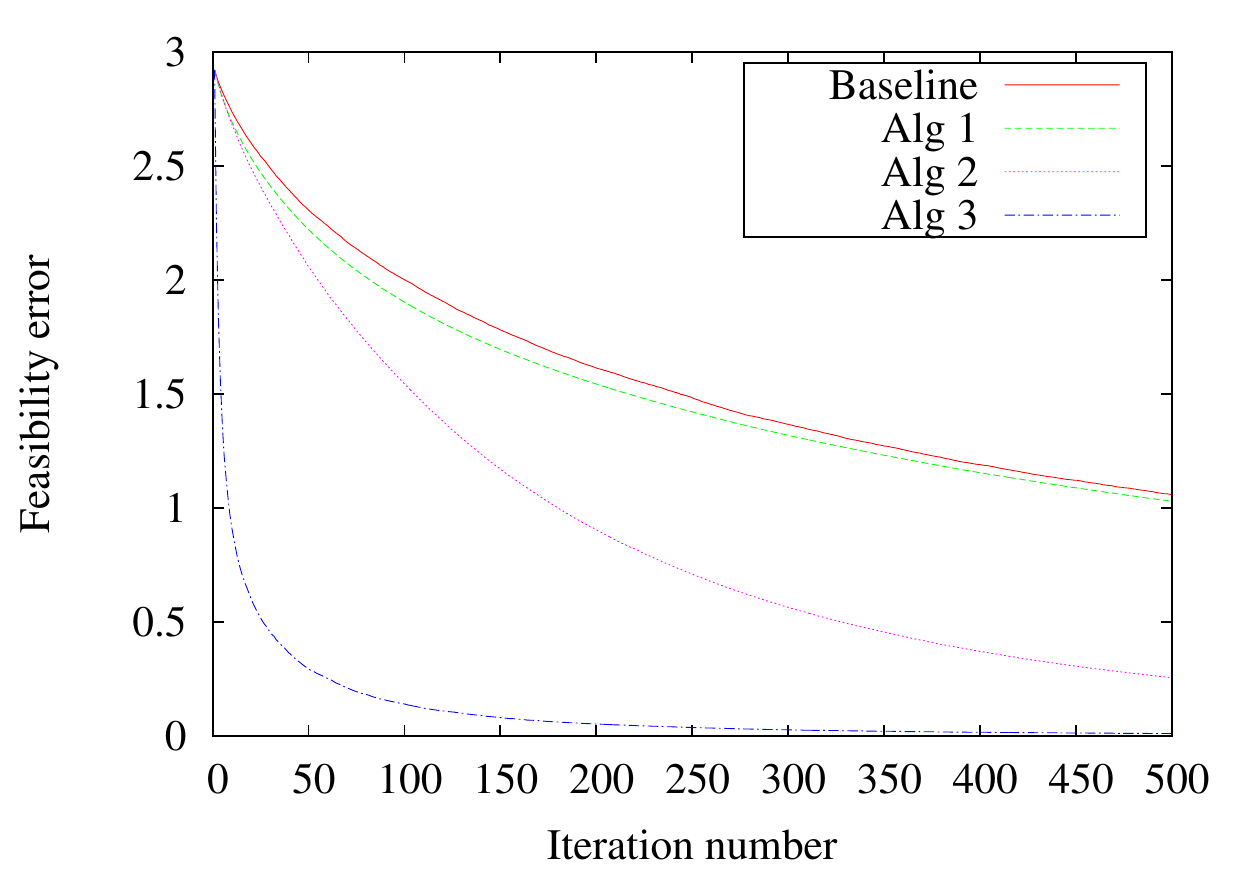}}&
        {\includegraphics[width=0.33\textwidth]{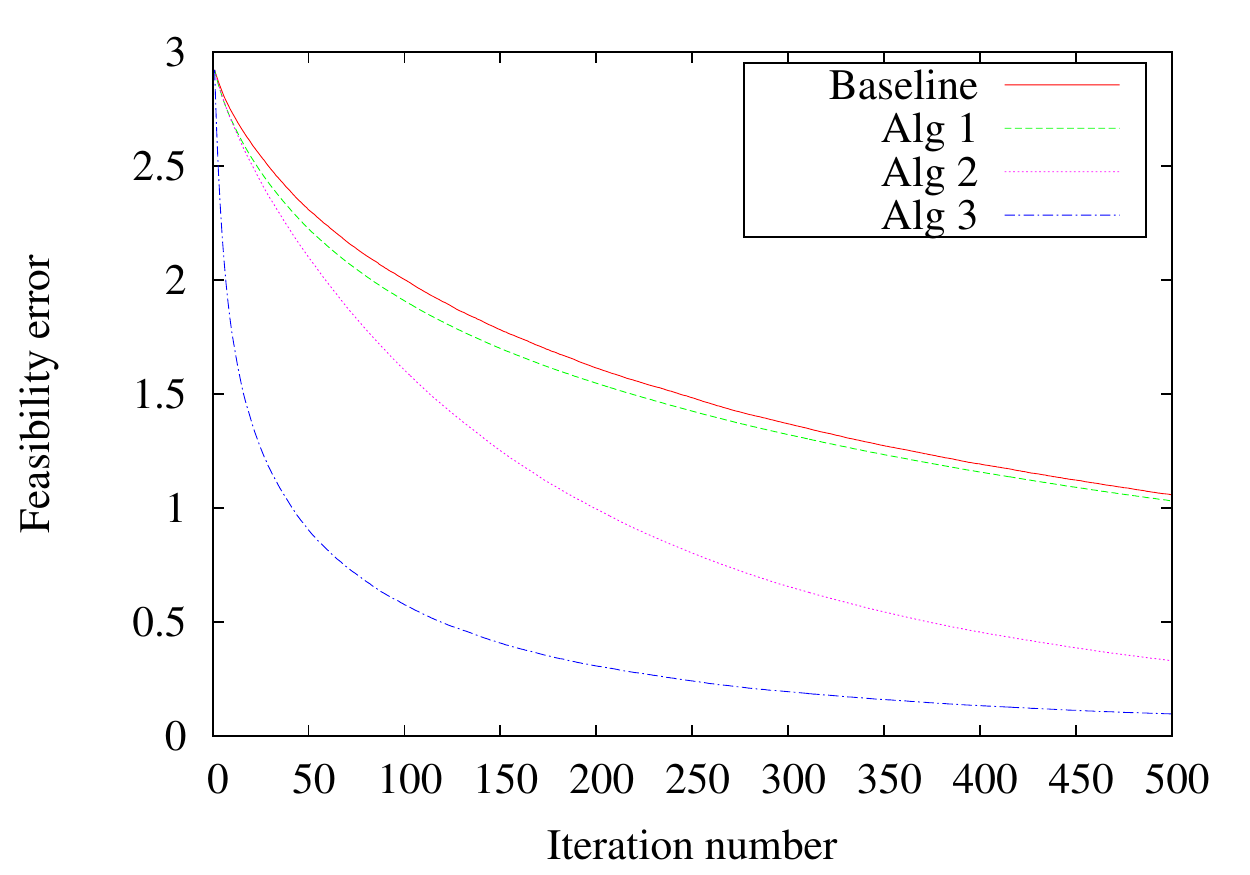}}&
        {\includegraphics[width=0.33\textwidth]{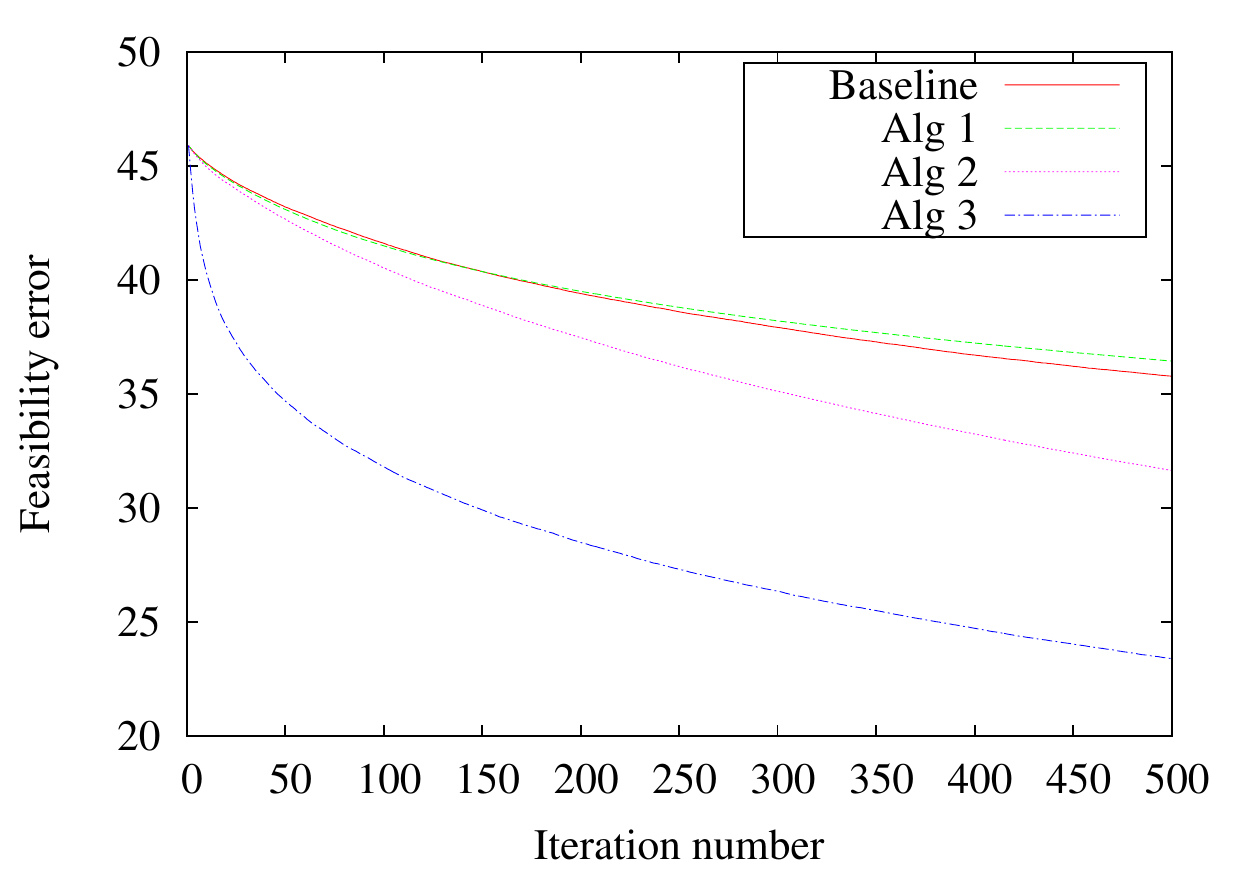}}\\
          {\small (a) $d=100, m=200, M=30$}&{\small (b) $d=100, m=200, M=10$}&{\small (c) $d=100, m=2000, M=10$}\\
    \end{tabular}
    \caption{Convergence of mean feasibility errors.} 
    \label{fig3}
    \end{figure}

\begin{figure}[htb]
    \begin{center}
    \begin{tabular}{ccc}
        {\includegraphics[width=0.33\textwidth]{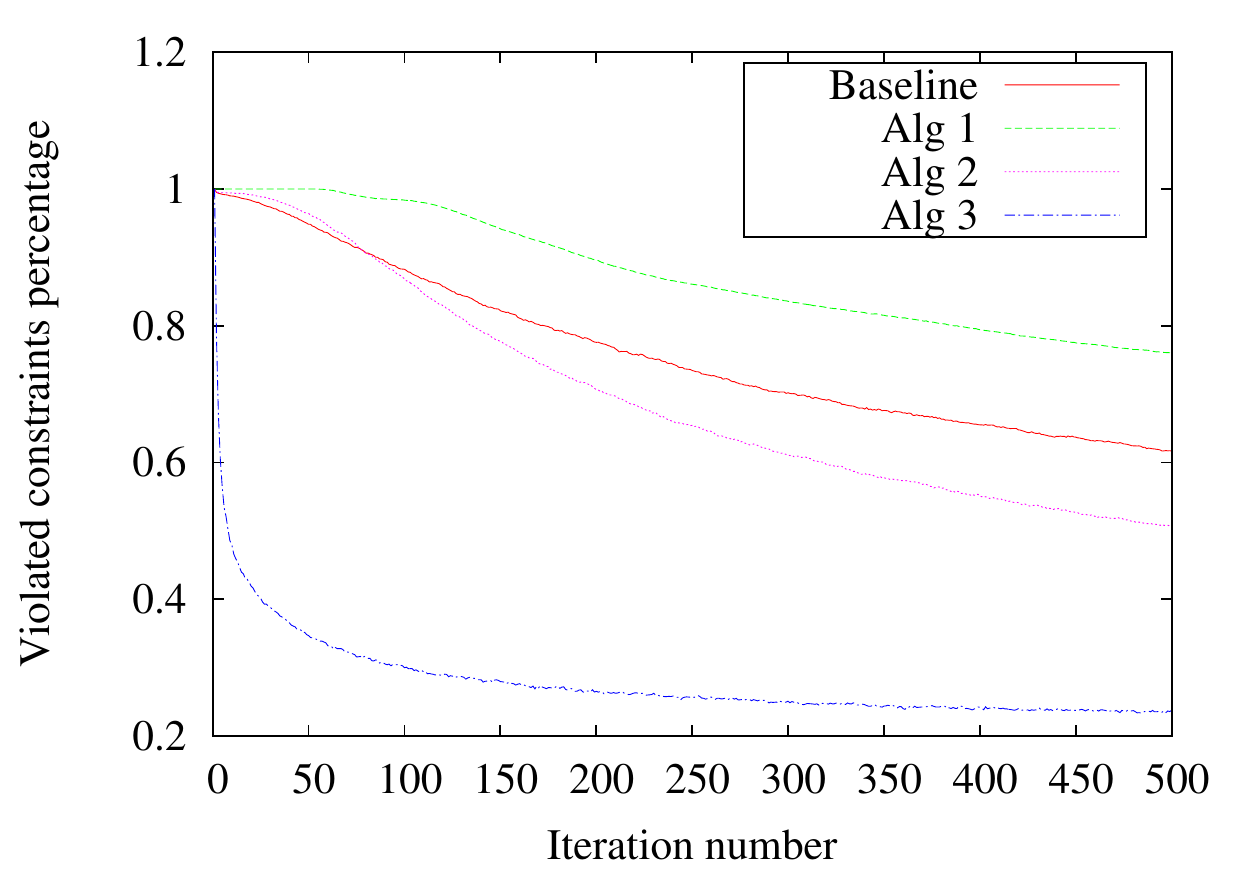}}&
        {\includegraphics[width=0.33\textwidth]{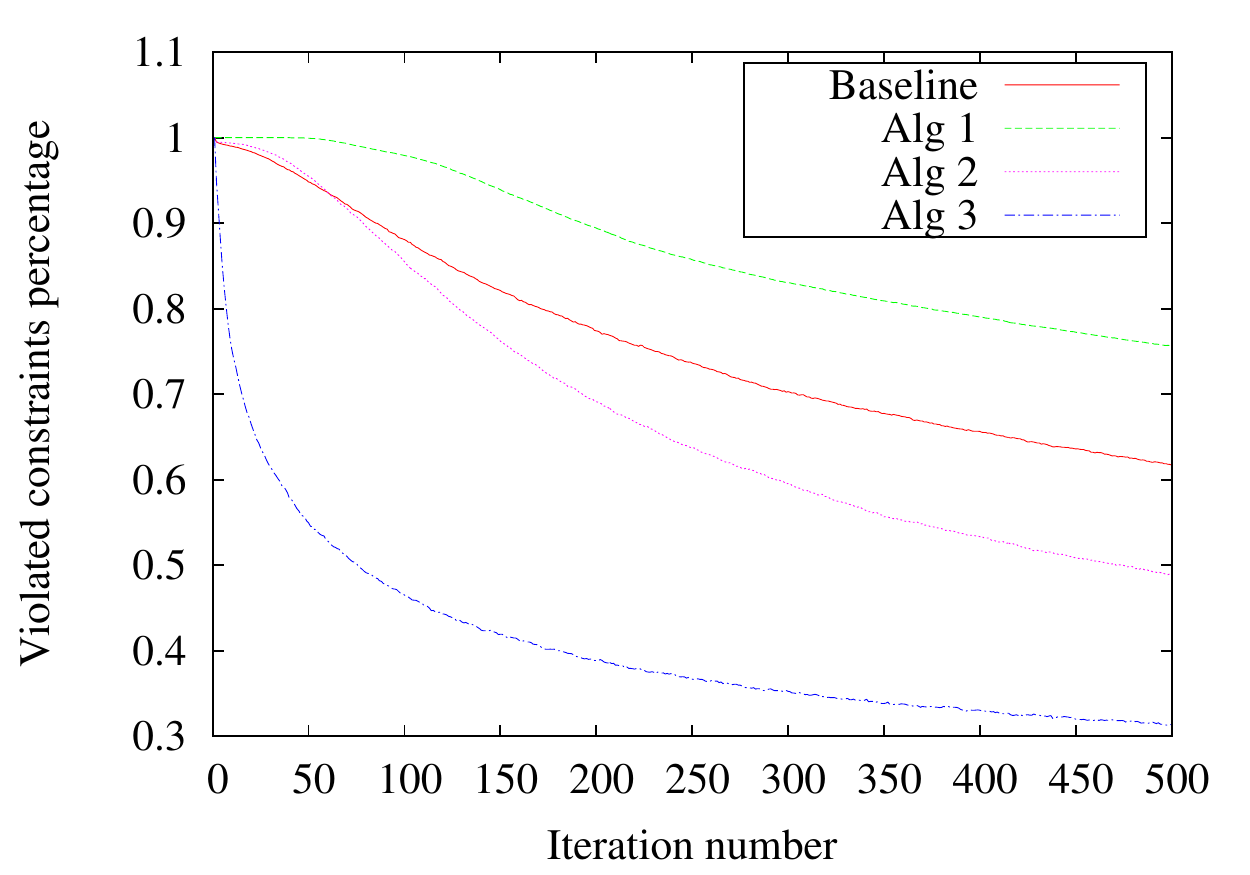}}&
        {\includegraphics[width=0.33\textwidth]{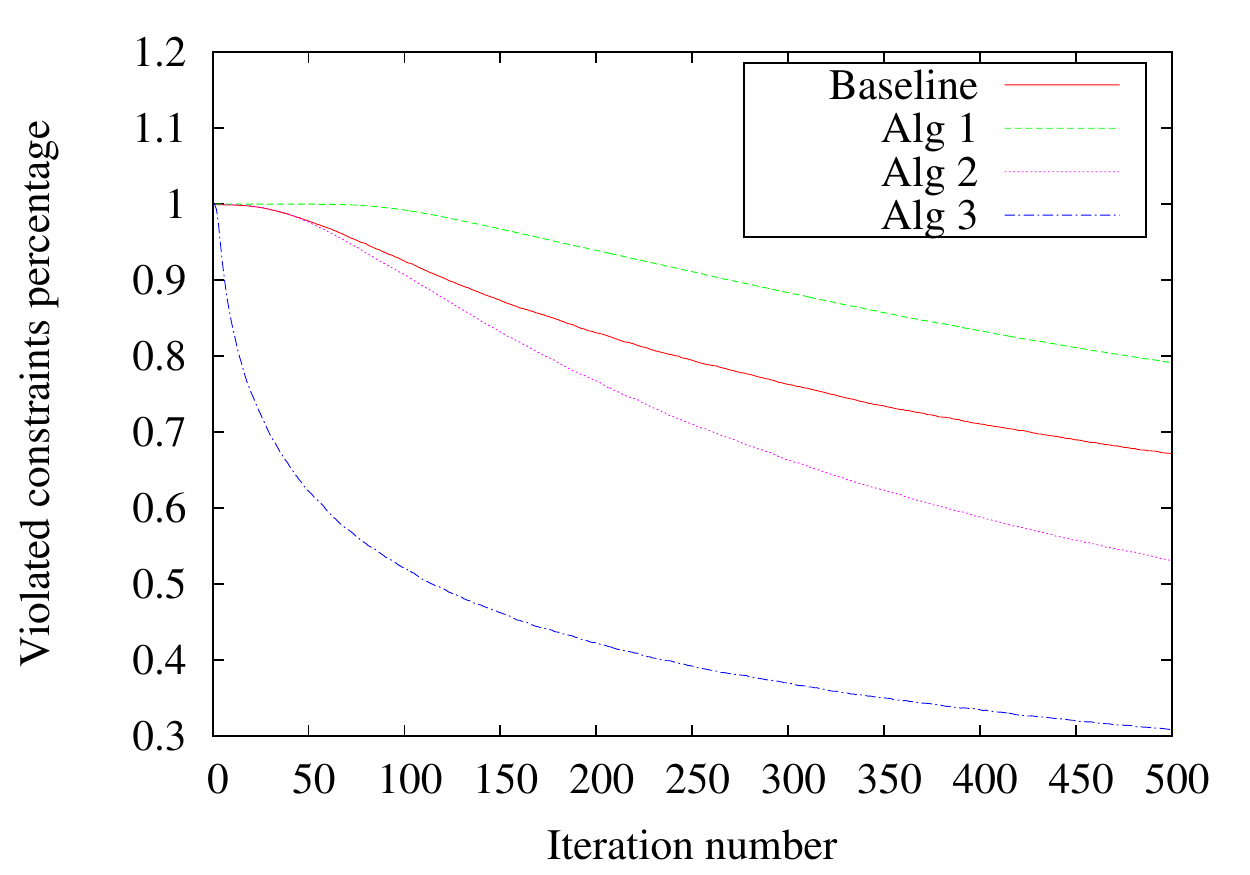}}\\
        {\small (a) $d=100, m=200, M=30$}&{\small (b) $d=100, m=200, M=10$}&{\small (c) $d=100, m=2000, M=10$}\\
    \end{tabular}
    \end{center}
    \caption{Percentage of constraint violation as the algorithms proceed.}
    \label{fig4}
\end{figure}

\begin{figure}[htb]
    \centering
    \begin{tabular}{cc}
        {\includegraphics[width=0.5\textwidth]{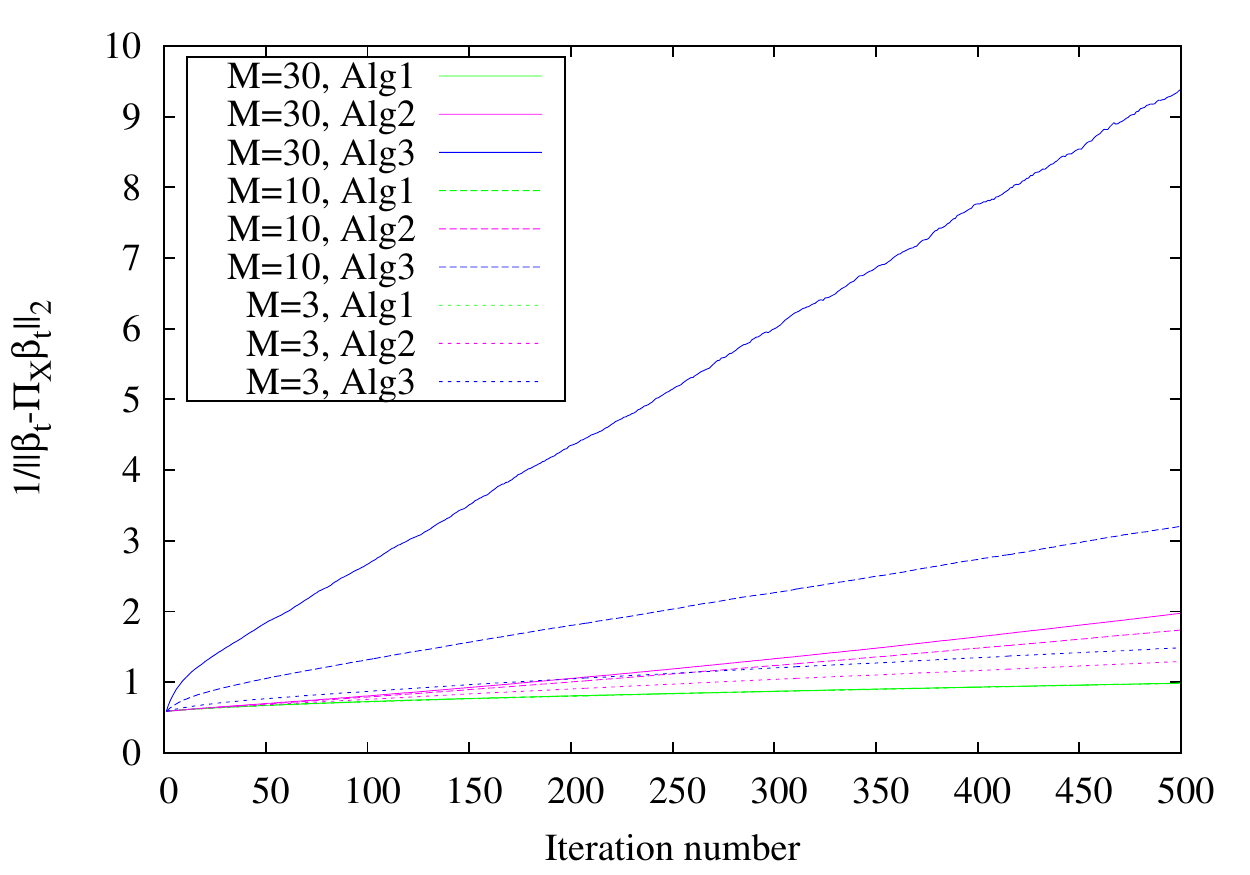}}&
        {\includegraphics[width=0.5\textwidth]{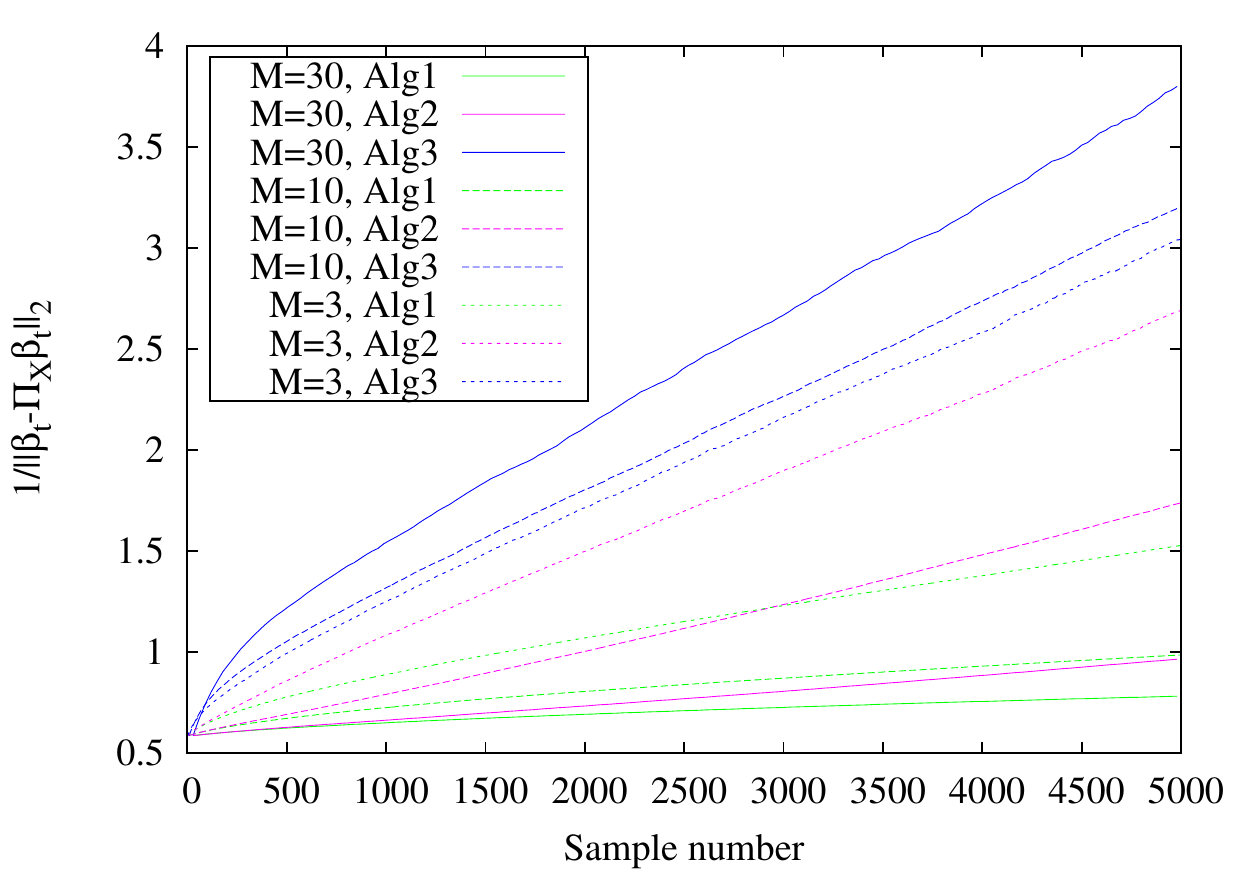}}\\
    \end{tabular}
    \caption{The inverse feasibility error is linearly related to the iteration number and the total constraint sample size. Left: The slope is a metric of iteration efficiency. Right: The slope is a metric of sample efficiency. }
    \label{fig5}
\end{figure}

We test Algorithms 1,2,3 and the baseline algorithm \eqref{alg-wmd} on instances of the support vector machine problem with varying dimension $d$ and data size (constraint size) $m$. We also test the algorithms with various values of $M$ (the number of sample constraints used per iteration). We let $\beta_0=0$ and let the stepsize be $1/(k+10)$ where $k$ is the iteration number. For each algorithm and each parameter setting, we generate $500$ trail runs. In Figure \ref{fig3} and Figure \ref{fig4}, we plot the convergence of mean feasibility error and the percentage of constraints violation, respectively.  Note that the mean optimality error is not reported here, as it is very similar to the mean feasibility error. In Figure \ref{fig5}, we illustrate the efficiency constants of various algorithms in terms of both the iteration efficiency and the sample efficiency.

According to Figure \ref{fig3}, when $M$ increases, the convergence rate of Algorithm 3 improves while that of Algorithm 1 and 2 remain largely unchanged. It hints that Algorithm 3 is more efficient in utilizing additional sample constraints. 
Figure \ref{fig4} illustrates the algorithms' ability to identify the feasible region.  From the figure, we can see that Algorithm 3 works far better than the other algorithms. Interestingly, we find that Algorithm 1 could be worse than the baseline algorithm.

Figure \ref{fig5} highlights the efficiency factors of various algorithms. It shows that the inverse feasibility error is linear to the iteration number and the total sample budget.  It verifies our theory that the feasibility error $\|\beta_t-\Pi_{\cX}\beta_t\|_2^2$ decreases on the order of $\mathcal{O}\left(\frac1{k^2}\right)$ (Theorem 3). The slopes in Figure \ref{fig5} correspond to the constant hidden in the error bounds, therefore they are metrics of efficiency for the tested algorithms.

The right plot of Figure \ref{fig5} is particularly interesting.  It plots the inverse error against the total number of sample constraints used so far. It shows that Algorithm 3 works uniformly better than Algorithm 2, and that Algorithm 2 works uniformly better than Algorithm 1, when the total sample size is fixed (even if $M$ varies). It indicates that  Algorithm 3 is more efficient in utilizing the samples and therefore has a better sample complexity (in addition to better iteration complexity). Let us focus on Algorithm 3 with different values of $M$, we observe that the sample efficiency increases as $M$ increases. In other words, by using $M$ sample constraints per iteration, the efficiency improvement is more than $M$ times for Algorithm 3. This verifies our conjecture in Section 4.3. It also suggests that using multiple constraints samples per iteration is more efficient than using one sample per iteration.

\section{Conclusion}
\label{sec_con}

We have proposed a class of random algorithms that involve stochastic (sub)gradients and random multi-constraint projections. We provide almost sure convergence and rate of convergence analysis for these algorithms. In particular, we have provided rate of convergence in terms of the optimality error and the feasibility error for a variety a feasibility update schemes, in the cases of general convex objectives and strongly convex objectives. See Table 1 for a summary of the convergence results.  These results suggest that, by using random projection in replacement of expensive exact projection, the stochastic gradient algorithm remain efficient with non-improvable iteration complexity (up to  a constant factor).

In comparison with the existing algorithm that uses one sample constraint per iteration, we show that using multiple sample constraints achieves significantly faster convergence rate. Within known algorithms, we show that the polyhedral-set projection algorithm achieves the best iteration complexity and sample complexity. 
Numerical experiments are provided to justify the theoretical results.

%

\bibliographystyle{alpha}
\bibliography{SCGD,SCGD1,vi,singular}

\end{document}